\documentclass[onefignum,onetabnum]{siamart171218}

\newcommand{\SM}{\M{K}}
\newcommand{\HM}{\M{H}} 
\newcommand{\DDS}{\M{S}} 
\newcommand{\DecP}{\tilde{\M{W}}\tilde{\M{W}}^*} 
\newcommand{\numv}{n}

\usepackage{amssymb,amsfonts,amsmath}

\usepackage[mathscr]{eucal}

\usepackage{stmaryrd}

\usepackage{amsbsy}

\usepackage{bm}

\usepackage{braket}

\usepackage{xparse}

\NewDocumentCommand \RangeSet { G{N} } {[\mathinner{#1}]}

\NewDocumentCommand \LinearSymbol {} {\oast}
\NewDocumentCommand \LeftSymbol {} {\blacktriangleleft}
\NewDocumentCommand \RightSymbol {} {\blacktriangleright}
\NewDocumentCommand \UnfoldSymbol {} {\oslash}

\NewDocumentCommand \SingleSize {s O{J} G{n}} {\IfBooleanTF{#1}{\bar}{} {#2}_{#3}}
\NewDocumentCommand \LinearSize {s O{J}} {\IfBooleanTF{#1}{\bar}{} #2^{\LinearSymbol}}
\NewDocumentCommand \LeftSize {s O{J} G{n}} {\IfBooleanTF{#1}{\bar}{} {#2}_{#3}^{\LeftSymbol}}
\NewDocumentCommand \RightSize {s O{J} G{n}} {\IfBooleanTF{#1}{\bar}{} {#2}_{#3}^{\RightSymbol}}
\NewDocumentCommand \UnfoldSize {s O{J} G{n}} {\IfBooleanTF{#1}{\bar}{} {#2}_{#3}^{\UnfoldSymbol}}
\NewDocumentCommand \FullSize{s O{J} G{N}} {%
  \IfBooleanTF{#1}%
  {\SingleSize*[#2]{0} \times \SingleSize*[#2]{1} \times \cdots \times \SingleSize*[#2]{#3-1}}%
  {\SingleSize[#2]{0} \times \SingleSize[#2]{1} \times \cdots \times \SingleSize[#2]{#3-1}}%
}

\NewDocumentCommand \SizeVec {s O{J}} {\mathbf{\IfBooleanTF{#1}{\bar}{} {#2}}}

\NewDocumentCommand \SingleIndex {s O{j} G{n}} {\IfBooleanTF{#1}{\bar}{} {#2}_{#3}}

\NewDocumentCommand \FullIndex{s O{j} G{N}} {%
  \IfBooleanTF{#1}%
  {(\SingleIndex*[#2]{0}, \SingleIndex*[#2]{1}, \dots, \SingleIndex*[#2]{#3-1})}%
  {(\SingleIndex[#2]{0}, \SingleIndex[#2]{1}, \dots, \SingleIndex[#2]{#3-1})}%
}

\NewDocumentCommand \IndexVec {s O{j}} {\mathbf{\IfBooleanTF{#1}{\bar}{} {#2}}}

\NewDocumentCommand \FullSubscript{s O{j} G{N}} {%
  \IfBooleanTF{#1}%
  {\SingleIndex*[#2]{0} \SingleIndex*[#2]{1} \dots \SingleIndex*[#2]{#3-1}}%
  {\SingleIndex[#2]{0} \SingleIndex[#2]{1} \dots \SingleIndex[#2]{#3-1}}%
}

\NewDocumentCommand \FacMat {s O{U} G{n}} {{\mathbf{\IfBooleanTF{#1}{\bar}{} #2}}^{(#3)}}
\NewDocumentCommand \FacSize {O{I} G{n}} {#1_{#2} \times R_{#2}}

\NewDocumentCommand \Tensor {s O{Y}} {\boldsymbol{\IfBooleanTF{#1}{\bar}{}
{\mathscr{\MakeUppercase{#2}}}}}
\NewDocumentCommand \ColumnBlock {O{\Mz{Y}{n}} G{\ell}} {#1 \left[#2\right]}

\newcommand{\Tra}{{\sf T}}

\newcommand{\V}[2][]{{\bm{#1\mathbf{\MakeLowercase{#2}}}}} 
 
\newcommand{\M}[2][]{{\bm{#1\mathbf{\MakeUppercase{#2}}}}} 
 
\newcommand{\Mz}[3][]{\M[#1]{#2}_{(#3)}}

\usepackage{subcaption}
\usepackage{booktabs}
\usepackage{adjustbox}
\usepackage{csvsimple}
\usepackage{bbm}
\usepackage{comment}

\newenvironment{varalgorithm}[1]
  {\algorithm\renewcommand{\thealgorithm}{#1}}
  {\endalgorithm}

\usepackage{lipsum}
\usepackage{amsfonts}
\usepackage{graphicx}
\usepackage{epstopdf}
\usepackage{algorithmic}
\ifpdf
  \DeclareGraphicsExtensions{.eps,.pdf,.png,.jpg}
\else
  \DeclareGraphicsExtensions{.eps}
\fi

\newsiamremark{remark}{Remark}
\newsiamremark{hypothesis}{Hypothesis}
\crefname{hypothesis}{Hypothesis}{Hypotheses}
\newsiamthm{claim}{Claim}

\headers{Skew-Symmetric Adjacency Matrices for Clustering Digraphs}{K. Hayashi, S. G. Aksoy, H. Park}

\title{Skew-Symmetric Adjacency Matrices for Clustering Directed Graphs\thanks{Under Review.
\funding{Koby Hayashi acknowledges support from the United States Department of Energy through the Computational Sciences Graduate Fellowship (DOE CSGF) under grant number: DE-SC0020347.}}}

\author{Koby Hayashi\thanks{School of Computational Science and Engineering, Georgia Institute of Technology, Atlanta, GA, USA
  (\email{khayashi9@gatech.edu}, \email{hpark@cc.gatech.edu}).}
\and Sinan G. Aksoy\thanks{Pacific Northwest National Laboratory, Seattle, WA, USA 
  (\email{sinan.aksoy@pnnl.gov})}
\and Haesun Park\footnotemark[2] }

\usepackage{amsopn}

\makeatletter
\newcommand*{\addFileDependency}[1]{
  \typeout{(#1)}
  \@addtofilelist{#1}
  \IfFileExists{#1}{}{\typeout{No file #1.}}
}
\makeatother

\newcommand*{\myexternaldocument}[1]{%
    \externaldocument{#1}%
    \addFileDependency{#1.tex}%
    \addFileDependency{#1.aux}%
}

\ifpdf
\hypersetup{
  pdftitle={Skew-Symmetric Adjacency Matrices for Clustering Directed Graphs},
  pdfauthor={K. Hayashi, S. G. Aksoy, H. Park}
}
\fi

\myexternaldocument{ex_supplement}

\begin{document}


\maketitle

\begin{abstract}
Cut-based directed graph (digraph) clustering often focuses on finding dense within-cluster or sparse between-cluster connections, similar to cut-based undirected graph clustering methods.
In contrast, for flow-based clusterings the edges between clusters tend to be oriented in one direction and have been found in migration data, food webs, and trade data.
In this paper we introduce a spectral algorithm for finding flow-based clusterings.
The proposed algorithm is based on recent work which uses complex-valued Hermitian matrices to represent digraphs.
By establishing an algebraic relationship between a complex-valued Hermitian representation and an associated real-valued, skew-symmetric matrix the proposed algorithm produces clusterings while remaining completely in the real field.
Our algorithm uses less memory and asymptotically less computation while provably preserving solution quality. 
We also show the algorithm can be easily implemented using standard computational building blocks, possesses better numerical properties, and loans itself to a natural interpretation via an objective function relaxation argument. 
\end{abstract}



\section{Introduction}
Many methods for undirected graph clustering focus on finding minimal cuts between dense clusters \cite{Luxburg2007} and many directed graph (digraph) clustering methods seek to extend this idea \cite{gleich2006hierarchical,digraph_clust_survey}.
However, there has also been attention paid to finding large `imbalanced cuts' in digraphs.
These are cuts where most of the edges are oriented from one cluster to the other with few oriented in the reverse direction.
Such cuts are present in migration data \cite{Cucuringu2019}, food webs \cite{JCN_BCS,pan_li_neurips}, and trade data \cite{Laenen_neruips}.
We refer to this dichotomy as density versus flow-based clustering.

Spectral approaches are frequently used for density-based graph clustering \cite{Luxburg2007,andrew_Ng_spec_clust,Vemplala_spec_clust,Peng_spec_clust}.
More recently, researchers have also applied spectral techniques for finding flow-based clusterings.
Spectral algorithms for mining flow-based patterns vary primarily in the matrix used to represent the digraph.
The matrix representations used can be broadly classified as general non-symmetric matrices, symmetrizations of the adjacency matrix, and complex-valued Hermitian matrices.
In particular, complex-valued Hermitian matrices have received much recent attention for encoding a wide variety of digraph structural properties. \cite{Laenen_neruips,Cucuringu2019,MOHAR_w6,furutani2019graph,zhang2021magnet}. 

Complex-valued Hermitian adjacency or Laplacian matrices are appealing because they have many nice theoretical properties.
Like real-valued symmetric matrices, complex-valued Hermitian matrices are subject to the spectral theorem, min-max theorem for eigenvalues, eigenvalue interlacing properties, and more.
In these complex-valued Hermitian matrix representations, complex numbers are used to encode edge direction. 
This is a potential advantage over symmetrization approaches which often lose information related to edge direction and asymmetric representations which often lack useful theoretical properties.
Some recent applications of Hermitian representations include the so-called magnetic digraph Laplacian's utilization in signal processing  \cite{furutani2019graph} and node classification and link prediction \cite{zhang2021magnet}. 
There has also been interest in developing a spectral theory for other, closely related complex-valued Hermitian matrices \cite{Guo_mixed_rep, liu2015hermitian} . 
For flow-based clustering, Cucuringu et al. \cite{Cucuringu2019} proposed an algorithm for finding imbalanced cuts based on a complex-valued Hermitian digraph adjacency matrix whose effectiveness they demonstrate via an analysis of a Directed Stochastic Block Model (DSBM) and empirical studies on real data.

In this work, we propose and compare spectral clustering algorithms for finding imbalanced cuts in digraphs.
The main results are facilitated by an algebraic relationship between Cucuringu et al.'s complex-valued matrix and an associated real-valued matrix, which we analyze via application of the Real Schur Decomposition.
This relationship enables a couple alternative algorithms to the one proposed in \cite{Cucuringu2019}.
Our proposed algorithms utilize asymptotically less memory and computation, while provably preserving solution quality.
Additionally, it can be easily implemented using standard computational building blocks, possesses better numerical properties, and loans itself to a natural interpretation via an objective function relaxation argument. 
Lastly, we empirically demonstrate these advantages on both synthetic and real world data.
In the later case it is demonstrated that the method can find meaningful flow-based cluster structures.

The paper is organized as follows: in Section \ref{sec:prelims} we establish notation, review relevant background, and review prior work on complex-valued digraph matrices. In Section \ref{sec:CAM_sec} the algorithm is derived using the aforementioned algebraic relationship and motivated using a heuristic relaxation argument which aides in the interpretability of the method. 
Finally, in Section \ref{sec:exp_sec} we present experimental results on the DSBM and Food Web data sets. 


\section{Preliminaries}\label{sec:prelims}
\paragraph{Definitions and notation}
We use $a \in \mathbb{C}$ to denote scalars, $\V{a} \in \mathbb{C}^{n}$ for vectors, $\M{A} \in \mathbb{C}^{m \times n}$ for matrices, $\mathscr{A}$ for sets, and the superscripts $\Tra$ and $*$ for transposition and conjugate transpose, respectively. A directed graph or digraph $G=(\mathscr{V},\mathscr{E})$ is a set of vertices $\mathscr{V}$ and a set $\mathscr{E}$ of {\it ordered} pairs of vertices, called edges.
Unless otherwise stated, we assume a digraph is accompanied by an edge-weighting function $w: \mathscr{E} \to \mathbb{R}_{\geq 0}$, and denote the weight of edge $(u,v)$ by $w_{uv}$. Further, if $(u,v) \in \mathscr{E}$, we sometimes write $u \rightarrow v$. If the digraph does not contain reciprocal edges, meaning $(u,v) \in \mathscr{E}$ implies that $(v,u) \not \in \mathscr{E}$, then we call the digraph an {\it oriented graph}. A $k$-partition of the vertex set of a graph is a set of non-empty, disjoint sets $\{\mathscr{A}_1, \cdots \mathscr{A}_k
\}$ such that $\bigcup_{j=1}^k \mathscr{A}_j = \mathscr{V}$.
The directed adjacency matrix associated with $G$ is $\M{M} \in \mathbb{R}^{\numv \times \numv}$, where $\M{M}_{uv} = w_{uv}$ if $u \rightarrow v \in \mathscr{E}$ and $0$ otherwise.
We frequently abuse notation by using vertex and cluster symbols as indices, e.g. $\M{M}_{uv} = w_{uv}$ as stated in the previous line.
Additionally, we use $n = |\mathscr{V}|$ and as a general positive integer, for example a square matrix $\M{A}\in \mathbb{C}^{n \times n}$.
For ease of reference and other notation, we provide Table \ref{tab:notation_table}.

\paragraph{Linear Algebra}
Some basic but relevant results in linear algebra, of which we make frequent use are as follows. 
A normal matrix $\M{B} \in \mathbb{C}^{m \times m}$ has an eigenvalue decomposition $\M{B} = \M{U}^{*}\M{\Lambda} \M{U}$, where $\M{\Lambda}\in \mathbb{C}^{m \times m} $ and diagonal and $\M{U} \in \mathbb{C}^{ m \times m }$ is unitary so $\M{U}\M{U}^* = \M{I}$. 
We will enforce the convention that $\M{\Lambda} = \mbox{diag}(\lambda_{1}, \cdots, \lambda_{m})$ where $|\lambda_i| \ge |\lambda_j|$ if $i \le j$.
All eigenvalues of a Hermitian matrix are real.
A matrix $\SM$ where $-\SM = \SM^\Tra$ is called skew-symmetric and real skew-symmetric if $\M{K}\in \mathbb{R}^{n \times n}$. 
If $\SM$ is real skew-symmetric all its eigenvalues are either 0 or purely imaginary of the form $\alpha i$ where $\alpha \in \mathbb{R}$.
If $\alpha i$ is a nonzero eigenvalue of $\SM$ then so is $-\alpha i$ and if $\V{x}$ is an eigenvector of $\SM$ so is $\bar{\V{x}}$.
The direct sum of a set of matrices is written as $\M{B} = \M{B}_1 \oplus \M{B}_2 \oplus \cdots \oplus \M{B}_m = \mbox{diag} (\M{B}_1 , \cdots , \M{B}_m)$, where $\M{B} \in \mathbb{C}^{n \times n}$, $\M{B}_j \in \mathbb{C}^{n_j \times n_j}$, and $\sum_{j=1}^m n_j = n$.

\begin{table}[]
\centering
\scalebox{0.9}{
\begin{tabular}{| c | c || c | c |}
\hline
\text{Symbols} & \text{Meaning} & \text{Symbols} & \text{Meaning} \\
\hline
$\mathscr{E}$ & Edge set & $\text{Re}(\cdot)$ & Real part \\
$\mathscr{V}$ & Vertex set & $\text{Im}(\cdot)$ &  Imaginary part\\
$\numv$ & Number of vertices & $\oplus$ & Direct matrix sum\\
$[\M{A}]_+$ & Proj. to nonnegative orthant & $\mathscr{A}$ & A set, Euler script\\
$i = \sqrt{-1}$ & Complex unit & $\M{A}$ & A matrix, bold-uppercase\\
$\M{E}^\Tra = 
\begin{bmatrix}
\M{I} & \M{0}
\end{bmatrix}$ & Truncation matrix & $\V{a}$ & A vector, bold-lowercase\\
$\mathbbm{1}$ & Vector of all ones & $|\cdot|$ & Absolute value or cardinality\\
$\M{A}^\Tra$ & Transposition & $\M{M}$ & Directed Graph Adj. \\
$\bar{\M{A}}$ & Complex conjugation & $\M{H}$ & Purely Hermitian Adj.\\
$\M{A}^* = \bar{\M{A}}^\Tra$ & Conjugate Transpose & $\M{K}$ & Real Skew-Symmetric Adj.\\
$\| \cdot \|_F$ & Frobenius norm & $\|\cdot \|_2$ & 2-norm\\
\hline
\end{tabular}
} 
\caption{Notation Table}
\label{tab:notation_table}
\end{table}

\subsection{Complex-valued digraph matrices}
Researchers have introduced a variety of different complex-valued adjacency matrices for studying digraphs, utilizing them for different purposes. 
Liu and Li \cite{liu2015hermitian} proposed a Hermitian adjacency matrix $\M{A}$ where $\M{A}_{uv} = 1$ if $(u,v) \in \mathscr{E}$ and $(v,u) \in \mathscr{E}$, $i$ if $(u,v) \in \mathscr{E}$ and $(v,u) \not \in \mathscr{E}$, $-i$ if $(u,v) \not \in \mathscr{E}$ and $(v,u) \in \mathscr{E}$ and 0 otherwise.

Their motivation for proposing this matrix is that it encodes the directionality of the digraph while possessing strictly real eigenvalues. This enabled a meaningful definition of {\it Hermitian energy} of digraphs for applications in theoretical chemistry for computing the $\pi$-electron energy of a conjugated carbon molecule \cite{liu2015hermitian}. Concurrently, Guo and Mohar \cite{Guo_mixed_rep} proposed the same matrix for the purposes of establishing a basic spectral theory of digraphs.
Later, Mohar \cite{MOHAR_w6} proposed a modification of the matrix, identically defined except that the complex unit $i$ is replaced with a sixth root of unity $\omega=(1+i\sqrt{3})/2$. $\omega$ is chosen due to the fact that $\omega \cdot \bar{\omega}= \omega+\bar{\omega}=1$, which ensures the matrix encodes combinatorial properties of digraphs, such as $\M{A}^k$ counting directed walks of length $k$. 
In Mohar's work and others, the choice of complex number is a parameter used to define the matrix, and is left to the user. For instance, in the $q$-adjacency matrix used to define magnetic Laplacian \cite{zhang2021magnet}, the parameter $q$ controls the choice of complex number in polar form. Taking $q=1/4$ and $q=1/6$ yields matrices almost identical to the aforementioned matrices, respectively, differing only in that non-reciprocal edges are weighted by a factor of $1/2$.

For this work, we focus on a related, but simplified complex-valued digraph adjacency matrix using the imaginary unit $i$ which is used by Cucuringu et al. \cite{Cucuringu2019}. This matrix $\M{H} \in \mathbb{C}^{\numv \times \numv}$, most properly defined for weighted, oriented digraphs, is given element-wise by 

\begin{equation}
\label{eq:comp_adj_def_weight} 
\M{H}_{uv} = \begin{cases} w_{uv} \cdot i & \mbox{ if } u \rightarrow v \\ -w_{vu}\cdot i & \mbox{ if } v \rightarrow u \\ 0 & \mbox{ otherwise} \end{cases}
\end{equation}
Clearly $\M{H}$ is Hermitian by definition. 
While defined for oriented graphs, this matrix can be naturally applied to general digraphs by replacing any pair of reciprocal edges between $u$ and $v$ with a single edge $(u,v)$ having weight $w_{uv}-w_{vu}$ if $w_{uv} \geq w_{vu}$.

\subsection{Imbalanced Cuts}
\label{sec:cucu_rev}
Cucuringu et al. \cite{Cucuringu2019} use the matrix $\M{H}$ as the adjacency matrix of an oriented graph.
Via a statistical argument based on a proposed Directed Stochastic Block Model (DSBM) the authors argue that the eigenvectors of $\M{H}$ can recover flow-based clusterings.
The proposed DSBM is designed such that digraphs generated from it are expected to have large `imbalanced cuts' between them.
That is, cuts where most of the edges are oriented from one cluster to the other and few in the reverse direction.
To measure the quality of a cut \cite{Cucuringu2019} uses a  quantity called the Cut Imbalance (CI), which is defined as:
\begin{equation}
    \label{eq_CI}
        \text{CI}(\mathscr{X},\mathscr{Y}) = \frac{w(\mathscr{X},\mathscr{Y})}{w(\mathscr{X},\mathscr{Y}) + w(\mathscr{Y},\mathscr{X})},
\end{equation}
where $\mathscr{X},\mathscr{Y}$ are sets of vertices such that $\mathscr{X} \cup \mathscr{Y} = \mathscr{V}$, $\mathscr{X} \cap \mathscr{Y} = \emptyset$, and $w(\mathscr{X},\mathscr{Y}) = \sum_{u \in \mathscr{X}, v \in \mathscr{Y}} \M{M}_{uv}$, i.e., the sum of edges oriented from $\mathscr{X}$ to $\mathscr{Y}$.
One can see that a large CI$(\mathscr{X},\mathscr{Y})$ value means that most of the edges are oriented from $\mathscr{X}$ to $\mathscr{Y}$, a small CI value means most edges are oriented from $\mathscr{Y}$ to $\mathscr{X}$ and a CI value close to $\frac{1}{2}$ means that the cut is balanced in the sense that $w(\mathscr{X},\mathscr{Y}) - w(\mathscr{Y},\mathscr{X})$ is close to 0.
A number of extensions to the multi-cluster case are also presented in \cite{Cucuringu2019}.

In order to make this a maximization problem, the equation $|\text{CI}(\mathscr{X},\mathscr{Y}) - \frac{1}{2}|$ is considered instead.
Since the goal over each of these metrics is to find a maximum over the partition. 
This problem is different from most cut based graph clustering techniques that seek to minimize the measure, such as the normalized or ratio cut which are common objective functions used to motivate standard spectral clustering \cite{normalized_cuts_image_seg,Luxburg2007}, as the goal is to find a partition that minimizes such quantities.

Cucuringu et al.'s statistical analysis of the DSBM bounds, with a certain probability, the number of vertices misclassified by their spectral algorithm.
As previously mentioned, this algorithm uses the eigenvectors corresponding to the largest magnitude eigenvalues of $\M{H}$. 
However, Cucuringu et al.~provide no direct connection between the CI, Eqn. \ref{eq_CI}, and their spectral algorithm.
Algorithm \ref{alg:herm_clust} presents details of this spectral algorithm, which we refer to as Hermitian Clustering (Herm).

\begin{varalgorithm}{Herm}
    \caption{Hermitian Clustering (Herm)} 
    \label{alg:herm_clust}
    \begin{algorithmic}
     \STATE{\textbf{input}: A directed graph and desired number of clusters $k$.}
     \STATE{Construct $\HM \in \mathbb{C}^{\numv \times \numv}$ as described by Eqn. \ref{eq:comp_adj_def_weight}}
     \STATE{Assign $l = k$ if $k$ is even and $l = k-1$ if $k$ is odd}
     \STATE{Compute the $l$ largest magnitude eigenvalues and their corresponding eigenvectors of the matrix $\HM$ $\{ (\lambda_1,\V{w}_1), \cdots, (\lambda_l,\V{w}_l) \}$}
     \STATE{Compute $\M{P} = \sum_{j=1}^{l} \V{w}_j \V{w}_j^* \in \mathbb{R}^{\numv \times \numv}$}
     \STATE{Run k-means on $\numv$ rows of $\M{P}$ with $k$ clusters}
     \RETURN{$k$ vertex clusters}
    \end{algorithmic}
\end{varalgorithm}

\section{Proposed Algorithm}\label{sec:CAM_sec}
We now motivate and derive our proposed algorithm.
To this end, we first discuss aspects of Hermitian Clustering, highlighting properties of the matrix $\M{H}$ which the algorithm utilizes.
Then, we explore some matrix decompositions related to $\M{H}$.
From these matrix decompositions and their relationships, we derive a new, improved clustering algorithm and motivate its applicability to flow-based clustering using a relaxation argument.
\subsection{Motivation}
\label{SS_discussion}
Observe the matrix $\HM \in \mathbb{C}^{\numv \times \numv}$ utilized in Algorithm \ref{alg:herm_clust} is not only Hermitian but also skew-symmetric.
Further, given any digraph with adjacency matrix $\M{M} \in \mathbb{R}^{\numv \times \numv}$, $\HM$ may be written as $\HM = i\SM = i(\M{M} - \M{M}^\Tra)$ where $\SM = (\M{M} - \M{M}^\Tra) \in \mathbb{R}^{\numv \times \numv}$ is real skew-symmetric.
This implies that $\SM$ and $\HM$ have the same eigenvectors and there is a relationship between clustering based on $\HM$ or $\SM$.
Denote the Hermitian Eigenvalue Decomposition (EVD) of 
$\HM = \M{W} \M{\Lambda} \M{W}^{*}.$
Then the EVD of $\SM$ is
\begin{equation}
\SM = \M{W} (\bar{i}\M{\Lambda}) \M{W}^{*}.
\label{EVD_K}
\end{equation}
Note that while $\SM$ is a real valued matrix, its EVD requires complex valued matrices.

Algorithm \ref{alg:herm_clust} utilizes an even number of the leading eigenvectors of $\HM$ to form a low-rank representation of a graph.
Specifically, the matrix $\tilde{\M{W}} = [\V{w}_1, \cdots, \V{w}_l] \in \mathbb{C}^{\numv \times l}$ is formed, where $\V{w}_j$ is the $j$th column of $\M{W}$ and $l$ is an even, positive integer, then k-means with $k$ clusters is run on the product $\M{P} = \tilde{\M{W}}\tilde{\M{W}}^{*}$.
Due to the fact that if $(\lambda_j,\V{x}_j)$ is an eigenpair of $\SM$ so is $(-\lambda_j, \bar{\V{x}}_j)$, the matrix $\M{P} = \tilde{\M{W}}\tilde{\M{W}^{*}} \in \mathbb{R}^{\numv \times \numv}$ is real valued.
This means a standard $k$-means algorithm that takes real valued input can be run on $\M{P}$.

However, the formation of $\M{P} = \tilde{\M{W}}\tilde{\M{W}}^{*}$ may cause computational issues, despite having the desirable property of being real-valued.
First, the matrix $\M{P}$ is of size $\numv \times \numv$ which is as large as the input graph and is likely dense.
Therefore running k-means on and storing this $\M{P}$ can be prohibitively expensive for large problems.
Second, $\M{P}$ may incur additional numerical issues due to the formation of the product, and in fact may not be real valued \cite{GandVL}.
This can be overcome in a number of ways, for example by taking its real part and discarding the residual imaginary components.
In the next section we propose a solution to these problems by observing some algebraic relationships.

\subsection{Algebraic Properties}
\label{sec:Z_SVD}
The real, skew-symmetric matrix  $\SM \in \mathbb{R}^{\numv \times \numv}$ has a real-valued Singular Value Decomposition (SVD) of the form $\SM = \M{U}\M{\Sigma}\M{V}^\Tra$ where $\{ \M{U}, \M{\Sigma}, \M{V} \} \in \mathbb{R}^{\numv \times \numv}$.
Working with the SVD of $\SM$ is desirable due to the fact that computing it requires only real arithmetic (unlike the EVD) and reliable algorithms and software are readily available for its computation.
Here we will follow this idea of using the SVD in place of the EVD. 

The derivation of our algorithm relies on properties of the Real Schur Decomposition (RSD) of $\SM$ \cite{GandVL}.
Recall the Schur Decomposition (SD), as opposed to the RSD, decomposes an arbitrary matrix $\M{B} \in \mathbb{C}^{n \times n}$ into $\M{B} = \hat{\M{Q}}\M{R}\hat{\M{Q}}^*$ where $\hat{\M{Q}} \in \mathbb{C}^{n \times n}$, $\hat{\M{Q}}\hat{\M{Q}}^* = \M{I}$, and $\M{R} \in \mathbb{C}^{n \times n}$ is upper triangular.
We emphasize the EVD and SD are different in general.
Even for a real matrix, $\M{A} \in \mathbb{R}^{n \times n}$, its SD $\M{A} = \hat{\M{Q}}\M{R}\hat{\M{Q}}^*$ consists of a unitary matrix $\hat{\M{Q}} \in \mathbb{C}^{n \times n}$ and an upper triangular matrix $\M{R} \in \mathbb{C}^{n \times n}$ which is also complex in general.
Alternatively, the RSD of a matrix $\M{A} \in \mathbb{R}^{n \times n}$ is $\M{A} = \M{Q}\M{T}\M{Q}^\Tra$ where $\M{Q} \in \mathbb{R}^{n \times n}$, $\M{Q}^\Tra\M{Q} = \M{I}$, and $\M{T} \in \mathbb{R}^{n \times n}$, where 
$\M{T}$ is block upper triangular with either $2\times 2$ or $1 \times 1$ blocks on the diagonal,
instead of being upper triangular.

Returning now to $\SM$, since this matrix is real skew-symmetric, its RSD
\begin{equation}
\label{RSD_K}
\SM = \M{Q}\M{T}\M{Q}^\Tra
\end{equation}
has a special form \cite{greub1967linear},
where $\M{Q} \in \mathbb{R}^{\numv\times \numv}$, $\M{Q}^\Tra \M{Q} = \M{I}$ , 
and $\M{T}  = \M{T}_{1} \oplus \dots \oplus \M{T}_{b} \in \mathbb{R}^{\numv\times \numv}$
is block diagonal with diagonal blocks of size $1 \times 1$ or $2 \times 2$,
and $b$ is the total number of these blocks.
Since all nonzero eigenvalues of $\SM$ are purely imaginary, and appear in $\pm$ pairs,
one may assume that $\SM$ has $2s$ non-zero eigenvalues and
each $\pm$ pair of the $2s$ eigenvalues appears in a block $\M{T}_{j} \in \mathbb{R}^{2 \times 2}$ for $j=1,\dots,s$,
$\M{T}_{t} = 0 \in \mathbb{R}^{1 \times 1}$ for $t = (2s+1), \dots, b$.
We may further assume that each block $\M{T}_j$ has the form
$\M{T}_j = \begin{bmatrix} 0 \quad \alpha_j; -\alpha_j \quad 0 \end{bmatrix} $
whose eigenvalues are $\alpha_ji$ and $ -\alpha_ji$,
and the blocks $\M{T}_j$ are ordered in non-increasing order by $|\alpha_j|$ and $\alpha_j > 0$.
These real Schur vectors can be easily used to construct eigenvectors.
Observe that
$$\SM
\begin{bmatrix}
\V{q}_{2j-1} & \V{q}_{2j}
\end{bmatrix}
= \begin{bmatrix}
\V{q}_{2j-1} & \V{q}_{2j}
\end{bmatrix}
\begin{bmatrix}
    0 & \alpha_j \\
    -\alpha_j & 0 
\end{bmatrix} = \begin{bmatrix}
    -\alpha\V{q}_{2j}  & \alpha\V{q}_{2j-1}
\end{bmatrix},$$
and $\SM (\V{q}_{2j-1} + i\V{q}_{2j}) = -\alpha\V{q}_{2j} + i\alpha\V{q}_{2j-1} = i \alpha (\V{q}_{2j-1} + i\V{q}_{2j})$, so
$(\V{q}_{2j-1} + i\V{q}_{2j})$ is an eigenvector of $\SM$, and therefore also of $\HM$.

Generalizing this observation define $\M{J}  = \M{J}_{1} \oplus \dots \oplus \M{J}_{b} \in \mathbb{C}^{\numv\times \numv}$, 
where
${\M{J}}_j = 
\frac{1}{\sqrt{2}}\begin{bmatrix}
1 \quad -i ;
1 \quad i
\end{bmatrix} \in \mathbb{C}^{2 \times 2}$ 
for $j=1,\dots,s$,
and $\M{J}_{t} = 1 \in \mathbb{R}^{1 \times 1}$ for $t = (2s+1), \dots, b$.
Then $\M{J} \M{T} \M{J}^*  = \bar{i}\M{\Lambda}$,
since ${\M{J}}_j$, for $j=1,\dots,s$, unitarily diagonalizes $\M{T}_j$,
as ${\M{J}}_j \M{T}_j {\M{J}}_j^* = 
\begin{bmatrix}
-\alpha_j \bar{i} \quad 0 ;
0 \quad \alpha_j \bar{i}
\end{bmatrix}$.
Therefore, we have

\begin{equation}
    \label{eq:K_def}
    \SM = \M{Q}\M{T}\M{Q}^\Tra = (\M{Q}\M{J}^*)(\bar{i}\M{\Lambda})(\M{J}\M{Q}^\Tra)
     = \M{W} (\bar{i}\M{\Lambda}) \M{W}^*,
\end{equation}
an EVD of $\SM$ where $\M{W} = (\M{Q}\M{J}^*)$.
This shows a relationship between the eigenvectors $\M{W}$ in Eqn. \ref{EVD_K} 
and the real Schur vectors $\M{Q}$ in Eqn. \ref{RSD_K} of $\SM$.

We are now ready to discuss our first main proposition which enables our proposed algorithm.
\begin{proposition}
    \label{edm_prop}
    Let $\tilde{\M{Q}} = \M{Q}\M{E}  \in \mathbb{R}^{\numv \times l}$, where $\M{E} \in \mathbb{R}^{\numv \times l}$ is a truncation matrix 
whose $l$ columns are the first $l$ columns of the identity matrix of order $\numv$, $l$ is a positive even integer $\le 2s$. Assume that $\M{K}$ has $2s$ non-zero eigenvalues. Then the embedding $\tilde{\M{Q}}$ has the same Euclidean distances between all pairs of vertices as the embedding $\M{P} = \DecP \in \mathbb{R}^{\numv \times \numv}$.
\end{proposition}
\begin{proof}
Assume that a set of $d$ points are collected as the rows of a matrix $\M{Y} \in \mathbb{R}^{d \times f}$ as $\V{y}_1^T, \cdots, \V{y}_d^T$, where $f$ is the number of features. Then the Euclidean distance matrix \cite{EDM_Theory} of $\M{Y}$, $\Delta(\M{Y})$ whose $(i,j)$th element is the $L_2$-norm distance between row $i$ and row $j$,  $(\Delta(\M{Y}))_{ij} = \| \V{y}_i - \V{y}_j \|_2^2$, is defined as 
$$\Delta (\M{Y}) = \V{\mathbbm{1}} \ \text{diag}(\M{Y}\M{Y}^\Tra)^\Tra - 2\M{Y}\M{Y}^\Tra + \text{diag}(\M{Y}\M{Y}^\Tra)\V{\mathbbm{1}}^\Tra . $$
Since
\begin{align*}
    (\DecP(\DecP)^\Tra)^\Tra =  ({\DecP}\overline{\tilde{\M{W}}} \tilde{\M{W}}^\Tra) = (\DecP\overline{\DecP}) \\
    = (\DecP\DecP) = \DecP = \M{Q}\M{J}^*\M{E}(\M{Q}\M{J}^*\M{E})^* = \M{Q}\M{J}^*\M{E}\M{E}^\Tra\M{J}\M{Q}^\Tra = \tilde{\M{Q}} \tilde{\M{Q}}^\Tra,
\end{align*}
we have $\Delta(\DecP) = \Delta(\tilde{\M{Q}})$.
\end{proof}

Proposition \ref{edm_prop} implies an equivalent but simplified version of Algorithm \ref{alg:herm_clust}.
That is, compute the embedding $\tilde{\M{Q}}$ from $\SM$ and run k-means on it.
Since both embeddings have the same Euclidean distances between vertices it can be expected that the algorithms produce the same result.
A small issue that we now address is the use of the RSD.

\begin{proposition}
    \label{svd_prop}
    The embedding $\tilde{\M{Q}} = \M{Q}\M{E} \in \mathbb{R}^{\numv \times l}$ can be obtained from the Singular Value Decomposition of $\SM$.
\end{proposition}

\begin{proof}
Define the matrix $\M{Z} = \M{Z}_{1} \oplus \dots \oplus \M{Z}_{b}$, which has the same block structure as $\M{T}$, where
$ \M{Z}_{j} = [ 0 \quad -1 ; 1 \quad 0 ]$ for $j = 1, \cdots, s$
and $\M{Z}_{t} = 1$ for $t = (s+1), \cdots, b$.
Note that $\M{Z}$ is orthogonal.
Then 
\begin{equation}
\SM = \M{Q}\M{T}\M{Q}^\Tra =  \M{Q} (\M{T}\M{Z}) (\M{Z}^\Tra \M{Q}^\Tra),
\end{equation}
which is an SVD of $\SM$ where $\M{U} = \M{Q}$, $\M{\Sigma} = \M{T}\M{Z}$, and $\M{V} = \M{Q}\M{Z}$. The columns of $\tilde{\M{Q}}$ can be obtained from $\M{U}$ or $\M{V}$.
\end{proof}

As a result of Proposition \ref{svd_prop}, the embedding $\tilde{\M{Q}}$ can be easily obtained from computing the SVD of $\M{K}$ using a readily available high quality implementation. Algorithm \ref{alg:skew_clust} presents the psuedo code of our main algorithm, which we call Skew-Symmetric Clustering. 
In Skew-Symmetric Clustering, $k$-means is run on $\tilde{\M{Q}} \in \mathbb{R}^{\numv \times l}$ yielding a computational complexity of $O(t_{max}k^2  \numv)$, where $t_{max}$ is the maximum number of k-means iterations. In Hermitian Clustering, Algorithm \ref{alg:herm_clust}, k-means is run on $\M{P}$ for a computational complexity of $O(t_{max} k \numv^2)$.
Therefore Skew-Symmetric Clustering is asymptotically faster than Hermitian Clustering.
Empirical timing results are given in Section \ref{sec:exp_sec}.
For storage complexity Hermitian Clustering's dominating cost is storing $\M{P} \in \mathbb{R}^{\numv \times \numv}$ which requires $O(\numv^2)$ storage, whereas Skew-Symmetric Clustering has an equivalent or lower storage complexity of $O(\mbox{nnz}(\M{Z}) + \numv l)$, where $\mbox{nnz}(\M{Z})$ denotes the number of nonzero entries of $\M{Z}$.
Additionally, we note that only $l$ singular vectors are needed in Skew-Symmetric Clustering even though typically it takes $2l$ real dimensions to capture $l$ complex ones.

\paragraph{SVD-Search Algorithm}
\label{svd_search_sec}
Empirically we notice that for synthetic graphs the associated $\M{K}$ can have a variable number of outlying singular values.
Due to this we implement a variant of Skew-Symmetric Clustering where instead of taking the first $l$ singular vectors, where $l$ is some simple function based on the desired number of clusters $k$, the number of singular vectors taken is determined by a simple search heuristic.
A larger subset of singular values and vectors are computed and the `gap' is determined. 
Singular vectors above this gap are taken and used for the embedding.
Empirically we find that this method significantly outperforms Hermitian Clustering and standard Skew-Symmetric Clustering on the graphs generated by certain inputs for the DSBM.
Additionally, we experiment with manual inputs of the parameter $l$.
Results in Section \ref{sec:exp_sec} show that determining an appropriate $l$ has a large impact on the embedding quality.

\begin{varalgorithm}{Skew-F}
\caption{Skew-Symmetric Clustering (Skew-F)} 
\label{alg:skew_clust} 
\begin{algorithmic}
\STATE{\textbf{input}: A digraph adjacency matrix $\M{M} \in \mathbb{R}^{\numv\times \numv}$ and desired number of clusters $k$.}
 \STATE{Construct $\SM = \M{M} - \M{M}^\Tra$.}
 \STATE{Ensure the graph with adjacency $[\SM]_+$ is weakly connected.}
 \STATE{Let $l = k$ if $k$ is even and $l = k-1$ if $k$ is odd}
 \STATE{Compute a truncated SVD of $\SM = \tilde{\M{U}}\tilde{\M{\Sigma}}\tilde{\M{V}}^\Tra$ where $\{ \tilde{\M{U}},\tilde{\M{\Sigma}},\tilde{\M{V}} \} \in \mathbb{R}^{\numv 
 \times l}$}
 \STATE{Run k-means on $\tilde{\M{U}}$ for $k$ clusters}
 \RETURN{$k$ vertex clusters}
\end{algorithmic}
\end{varalgorithm}

\subsection{Trade Flow and Skew-Symmetric Clustering}
\label{sec:FI_SSC}

We now present a connection between our Skew-Symmetric Clustering algorithm and an intuitive cluster-quality metric called {\it Trade Flow}. Recall Eqn. \ref{eq_CI} defines a different metric, called Cut Imbalance (CI). Although \cite{Cucuringu2019} does not explicitly tie CI to Algorithm \ref{alg:herm_clust}, the authors utilize this metric to evaluate the cluster quality of their methods on real world datasets where no ground truth was available.
In place of the CI, we consider the Trade Flow (TF) metric for measuring imbalanced cuts, as proposed by Laenen \cite{laenen_MS_2019}:
\begin{equation}
    \label{eq_FI}
    \text{TF}(\mathscr{X},\mathscr{Y}) = |w(\mathscr{X},\mathscr{Y}) - w(\mathscr{Y},\mathscr{X})|.
\end{equation}
In the context of finding clusters with large imbalanced cuts the goal is to maximize $\text{TF}(\mathscr{X},\mathscr{Y})$ over the vertex clusters $\mathscr{X}$ and $\mathscr{Y}$, where $\mathscr{X}$ and $\mathscr{Y}$ form a partition.
Clearly, the TF is similar in spirit to the CI.
A large value of TF, Eqn. $\ref{eq_FI}$, means that more edge weight is oriented from one cluster to the other than vice versa.
A small value means that the cut is relatively balanced and thus by attempting to maximize Eqn. \ref{eq_FI} one expects to obtain clusters with large imbalanced cuts between them.

Now we present a heuristic, relaxation argument for why Skew-Symmetric Clustering can be expected to recover large imbalanced cuts.
Specifically, we will show that when $k=2$ our method can be viewed as maximizing Eqn. \ref{eq_FI} by relaxing the problem over the reals.
We note the TF problem for $k=2$ is solvable in linear time but for $k \ge 3$ is NP-hard \cite{laenen_MS_2019}. 
This relaxation is not meant as an improved algorithm but to connect the above methods to a reasonable objective function.

Consider two indicator vectors for the partition $\mathscr{X}$ and $\mathscr{Y}$ denoted $\V{e}_{\mathscr{X}}$ and $\V{e}_{\mathscr{Y}}$, where
$(\V{e}_{\mathscr{X}})_u = 1$ if vertex $u$ $\in \mathscr{X}$ and 0 otherwise.
Given a digraph we can write the TF in terms of the adjacency matrix $\M{M}$ as
\begin{align*}
    \text{TF}(\mathscr{X},\mathscr{Y}) = |w(\mathscr{X},\mathscr{Y}) - w(\mathscr{Y},\mathscr{X})|
    =  |\V{e}_{\mathscr{X}}^\Tra \M{M} \V{e}_{\mathscr{Y}} - \V{e}_{\mathscr{Y}}^\Tra \M{M} \V{e}_{\mathscr{X}}| \\ 
    = |\V{e}_{\mathscr{X}}^\Tra (\M{M} - \M{M}^\Tra) \V{e}_{\mathscr{Y}}| 
    = |\V{e}_{\mathscr{X}}^\Tra \SM \V{e}_{\mathscr{Y}}|
\end{align*}
Using the above observation we can then write the TF maximization problem as
\begin{equation}
    \label{FI_maximization}
    \max_{\mathscr{X},\mathscr{Y}} \text{TF}(\mathscr{X},\mathscr{Y}) = \max |\V{e}_{\mathscr{X}}^\Tra \SM \V{e}_{\mathscr{Y}}| \ \text{such that} \ \ \{ \V{e}_{\mathscr{X}},\V{e}_{\mathscr{Y}} \} \in \{0,1\}^{\numv} \ \text{and} \ \V{e}_{\mathscr{X}}^\Tra \V{e}_{\mathscr{Y}} = 0
\end{equation}
Relaxing this problem by allowing $\V{e}_{\mathscr{X}}$ and $\V{e}_{\mathscr{Y}}$ to take on arbitrary real values we can instead consider the problem as
\begin{equation}
\label{relaxed_prob}
\max |\V{a}_{\mathscr{X}}^\Tra \SM \V{b}_{\mathscr{Y}}| \ \text{such that} \ \| \V{a}_{\mathscr{X}} \| = \| \V{b}_{\mathscr{Y}} \| = 1,
\{ \V{a}_{\mathscr{X}}, \V{b}_{\mathscr{Y}}\} \in \mathbb{R}^{\numv}, \ \text{and} \ \V{a}_{\mathscr{X}}^\Tra\V{b}_{\mathscr{Y}} = 0
\end{equation}
Where the norm constraint deals with scaling and the orthogonality constraint takes the place of $\mathscr{X} \cap \mathscr{Y} = \emptyset$.

\begin{proposition}
\label{relax_prop}
Skew-Symmetric Clustering solves Equation \ref{relaxed_prob}. Therefore Skew-Symmetric Clustering can be viewed as solving a relaxation of Equation \ref{FI_maximization}, the Trade Flow maximization problem.
\end{proposition}

\begin{proof}
Consider the maximization problem in Eqn. \ref{relaxed_prob}, it is well known for an arbitrary matrix $\M{B}$ the quantity $\V{x}^\Tra\M{B}\V{y}$ is maximized by setting $\V{x}$ to equal the first left singular vector of $\M{B}$  and $\V{y}$ to equal the first right singular vector of $\M{B}$, assuming $\|\V{x}\| = \|\V{y}\| = 1$.
This is exactly what Skew-Symmetric Clustering does.
Additionally since $\SM$ is a real valued matrix its singular vectors are chosen to be real.

Next consider the orthogonality constraint, $\V{a}_{\mathscr{X}}^\Tra \V{b}_{\mathscr{Y}} = 0$.
In general one does not expect the first left and right singular vectors of a matrix to be orthogonal.
From Section \ref{sec:Z_SVD} we know one can write $\SM = \M{Q}\M{T}\M{Q}^\Tra =  \M{Q}(\M{T}\M{Z}) (\M{Z}^\Tra \M{Q}^\Tra)$ using its RSD.
As previously discussed this can be viewed as an SVD of $\SM$ where $\SM = \M{U}\M{\Sigma}\M{V}^\Tra$ where $\M{U} = \M{Q}$, $\M{V} = \M{Q}\M{Z}$ and $\M{T}\M{Z} = \M{\Sigma}$. 
It then follows that $\M{V}^\Tra \M{U} = \M{Z}^\Tra \M{Q}^\Tra \M{Q} = \M{Z}^\Tra$ and $\V{e}_1^\Tra\M{U}^\Tra \M{V}\V{e}_1 = \V{u}_1^\Tra \V{v}_1 = \V{e}_1^\Tra\M{Z}\V{e}_1 = 0$.
Therefore the first left and right singular vectors of $\SM$ are orthogonal.
\end{proof}

\section{Experiments}
\label{sec:exp_sec}
We now examine the empirical performance of our algorithms versus existing methods.
First, we demonstrate our methods performance on synthetic data sets generated from the DSBM.
In particular, we consider three different ways of generating the DSBM and include thorough experimental results for each.
Second, we explore our methods effectiveness when applied to real world data. We consider the following algorithms:
\begin{enumerate}
    \item Hermitian Clustering (Herm) see Algorithm \ref{alg:herm_clust}.
    \item Skew-Symmetric Clustering Full (Skew-F) see Algorithm \ref{alg:skew_clust}.
    \item Skew-Symmetric Clustering Reduced (Skew-R) which is the same as Skew-F but takes in a user specified parameter $l$.
    \item Skew-Symmetric Clustering Search (Skew-S) as in Algorithm \ref{alg:skew_clust} but modified as described in Section \ref{svd_search_sec}. That is the gap in the singular values is used to determine $l$.
    \item \ref{alg:biblio_clustering} computes $\DDS = \M{M}\M{M}^\Tra + \M{M}^\Tra \M{M}$ and uses the top $k$-eigenvectors to cluster via k-means \cite{DD_sym_citation}. It was one of the top performing algorithms compared against \ref{alg:herm_clust} in \cite{Cucuringu2019}.
    \item Block Cyclic Clustering (\ref{alg:bcs_clustering}) uses elements of Perron Frobenius theory to compute a vertex embedding from the row normalized adjacency matrix \cite{JCN_BCS}.
    \item \ref{alg:svd_clustering} computes $d$ left and right singular vectors of the adjacency matrix $\M{M}$, forms them into an embedding, and applies K-means to extract clusters \cite{svdm_paper}. We set $d=k$.
\end{enumerate}

\begin{varalgorithm}{BCS}
 \caption{Block Cyclic Spectral Clustering (BCS)} 
 \label{alg:bcs_clustering} 
\begin{algorithmic}
\STATE{\textbf{input}: A directed, strongly connected graph adjacency matrix $\M{M} \in \mathbb{R}^{\numv \times \numv}$ and desired number of clusters $k$.}
 \STATE{Construct $\M{P} = \M{D}_{out}^{-1}\M{M}$ where $\M{D}_{out} = \text{diag}(M)\V{\mathbbm{1}}$}
 \STATE{Compute $l = \lfloor \frac{k}{2} \rfloor$}
 \STATE{Compute the $l$ largest eigenvalues $\lambda_1,\cdots,\lambda_l$ of $\M{P}$ with largest absolute value that satisfy $\lambda \in \mathbb{C}: \text{Re}(\lambda) < 1, \text{Im}(\lambda) \ge 0$ and the associated right eigenvectors $\V{u}_1,\cdots,\V{u}_l$}
 \STATE{Collect the vectors into the matrix $\M{\Gamma} = [\V{u}_1,\cdots,\V{u}_l] \in \mathbb{C}^{n \times l}$}
 \STATE{Run k-means for $k$ clusters on the matrix $[\text{Re}(\M{\Gamma}), \text{Im}(\M{\Gamma})]$}
 \RETURN{$k$ vertex clusters}
\end{algorithmic}
\end{varalgorithm}
We note there are also normalized variants of the above algorithms, which we utilize later in Section \ref{sec:exp_real}. 
For completeness, we also note Laenen and Sun \cite{Laenen_neruips} give an algorithm for the circulant case of the DSBM, see Section \ref{sec:circ_dsbm}.
All algorithms we consider are generally applicable and not restricted to the circulant case.

\begin{varalgorithm}{SVD-M}
 \caption{SVD Clustering (SVD-M)} 
 \label{alg:svd_clustering} 
\begin{algorithmic}
\STATE{\textbf{input}: A directed graph adjacency matrix $\M{M} \in \{0,1\}^{\numv\times \numv}$, desired number of clusters $k$, and $d \in \{1,2,\cdots, \numv\}$}.
 \STATE{Compute the d-truncated SVD of $\M{M} \approx \hat{\M{U}}\hat{\M{\Sigma}} \hat{\M{V}}^\Tra$ where $\hat{\M{U}} \in \mathbb{R}^{\numv \times d}$, $\hat{\M{V}} \in \mathbb{R}^{\numv \times d}$, $\hat{\M{\Sigma}} \in \mathbb{R}^{d \times d}$}
 \STATE{Form $\tilde{\M{Z}} = [\hat{\M{U}}\hat{\M{\Sigma}}^{1/2}, \hat{\M{V}}\hat{\M{\Sigma}}^{1/2}]$}
 \STATE{Run k-means for $k$ clusters on the matrix $\tilde{\M{Z}}$}
 \RETURN{$k$ vertex clusters}
\end{algorithmic}
\end{varalgorithm}
\begin{varalgorithm}{DD-Sym}
 \caption{Bibliometric Clustering (DD-Sym)} 
 \label{alg:biblio_clustering} 
\begin{algorithmic}
\STATE{\textbf{input}: A directed graph adjacency matrix $\M{M} \in \{0,1\}^{\numv\times \numv}$, desired number of clusters $k$, and $0 \le \alpha \le 1$}
 \STATE{Compute $\M{A} = \alpha\M{M}\M{M}^\Tra + (1-\alpha)\M{M}^\Tra\M{M}$} 
 \STATE{Compute the first $k$ leading eigenvectors of $\M{A}$ and collect them in the matrix $\M{B}$}
 \STATE{Run k-means for $k$ clusters on the matrix $\M{B}$}
 \RETURN{$k$ vertex clusters}
\end{algorithmic}
\end{varalgorithm}

\paragraph{Cut Metrics}
To evaluate cluster quality, we utilize several multi-way cut objectives suitable for partitions with $k \ge 2$ clusters. 
First, we use a natural extension of the CI metric (Eqn. \ref{eq_CI}), defined in \cite{Cucuringu2019} as:
\begin{equation}
    \label{eq:CI_vol}
    \text{TopCI}^{vol}(\mathscr{A}_1,\cdots,\mathscr{A}_k) = \sum_{t=1}^{c} \text{CI}^{vol} (\mathscr{A}_{j_t},\mathscr{A}_{h_t})
\end{equation}
where $\text{CI}^{vol}(\mathscr{A}_{j_t},\mathscr{A}_{h_t})$ is the $t$-th largest $\text{CI}^{vol}$ pair of clusters, $\text{CI}^{vol}(\mathscr{X},\mathscr{Y}) = |\text{CI}(\mathscr{X},\mathscr{Y}) - 0.5|*\min(vol(\mathscr{X}),vol(\mathscr{Y}))$, and $vol(\mathscr{X})$ is the sum of all in and out degrees of vertices in $\mathscr{X}$.
A related metric, $\text{TopCI}^{sz}$ is defined by using the cardinality of a cluster in place of its volume in Eqn. \ref{eq:CI_vol}.
Secondly, for the TF metric (Eqn. \ref{eq_FI}) with $k\ge 2$, we have
\begin{equation}
    \label{eq:toptf}
    \text{TopTF}(\mathscr{A}_1,\cdots,\mathscr{A}_k) = \sum_{t=1}^{c} \text{TF} (\mathscr{A}_{j_t},\mathscr{A}_{h_t}) \ \text{such that} \ j_t \ge h_t,
\end{equation}
where, similar to the above, $\text{TF} (\mathscr{A}_{j_t},\mathscr{A}_{h_t})$ is the $t$-th largest TF pair of clusters. 
Note that an ordering over the indices is enforced due to the fact that the TF is symmetric in its inputs while CI is not.
In both equations $c$ is the number of cuts considered.
In contrast to \cite{Cucuringu2019} which fixes this parameter at $c=2k$, we vary it according to the problem structure. 

\subsection{Directed Stochastic Block Model Experiments}
\label{sec:exp_dsbm}
In our first set of experiments, we utilize the Directed Stochastic Block Model (DSBM) proposed in \cite{Cucuringu2019}. The DSBM is based on the inputs $(k,p,q,\V{c},\M{F})$, where
$k$ denotes the number of clusters, $p$ the probability that two vertices in the same cluster have an edge between them, $q$ the probability that two vertices in different clusters have an edge between them, $\V{c}$ a vector of length $k$ whose entries are the number of vertices in each cluster, and the matrix $\M{F} \in \mathbb{R}^{k \times k}$ which gives cluster level orientation probabilities.
That is if $u$ is in cluster $a$ and $v$ is in cluster $b$ then $u \rightarrow v$ exists with probability $\M{F}_{a,b}$.
All diagonal entries of $\M{F}$ are equal to $\frac{1}{2}$.
The graph corresponding to the entries of $\M{F}$ is called the {\it meta-graph} of a graph generated from this DSBM.


We utilize the following parameter settings: following \cite{Cucuringu2019} we set $p=q$ so that only the number of edges between clusters is expected to contain meaningful statistical information about the cluster memberships.
We vary $p=0.0045,0.008$, set $\numv = 5000$, $k = 5$, and assign each cluster 1000 vertices. Further, we vary a noise parameter $0 \le \mu < 0.5$ controlling the difficulty of recovery: if cluster $i$ is oriented to cluster $j$, then $\M{F}_{ij} = 1-\mu$ and $\M{F}_{ji} = \mu$. In this way, as $\mu$ approaches 0.5 the number of edges between clusters becomes random in expectation, making cluster recovery increasingly difficult; we consider 11 different $\mu$ values ranging from $0$ to $0.3$. We evaluate cluster quality using both the Adjusted Rand Index (ARI) \cite{ARI_cit} and TopTF (Eqn. \ref{eq:toptf}), with an appropriate value of $c$. Under this setup, we consider three versions of DSBM, differing with regard to the structure of $\M{F}$.

\begin{figure}
\centering
    \includegraphics[width=1\textwidth]{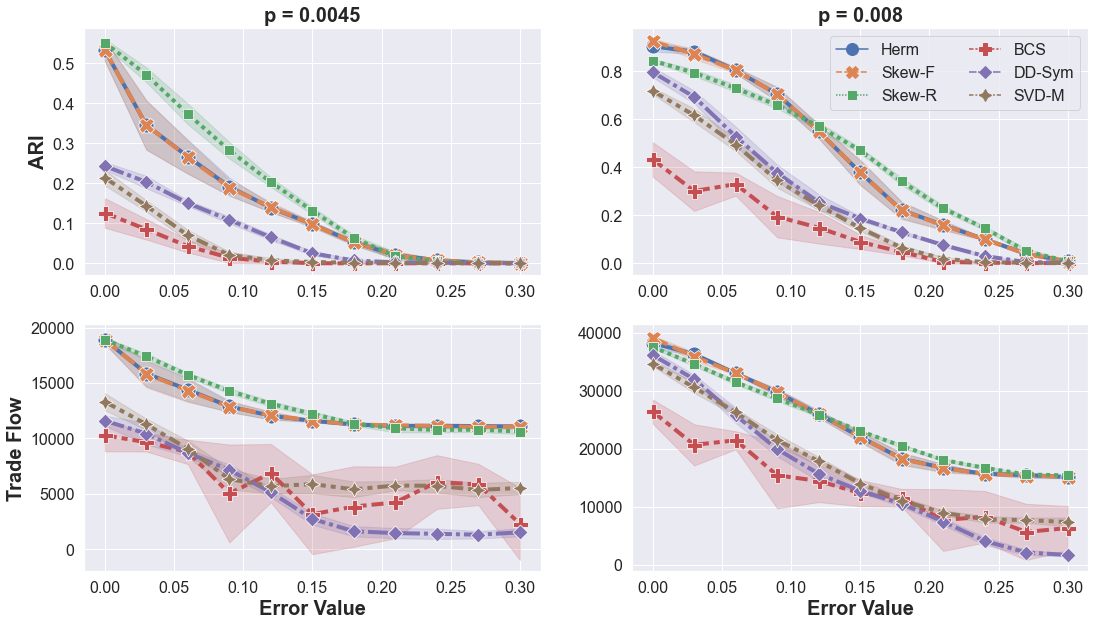}
    \caption{ARI and TopTF results for the Circulant DSBM experiments.}
  \label{fig:ari_cyclic}
\end{figure}

\begin{figure}
\centering
\includegraphics[width=1\textwidth]{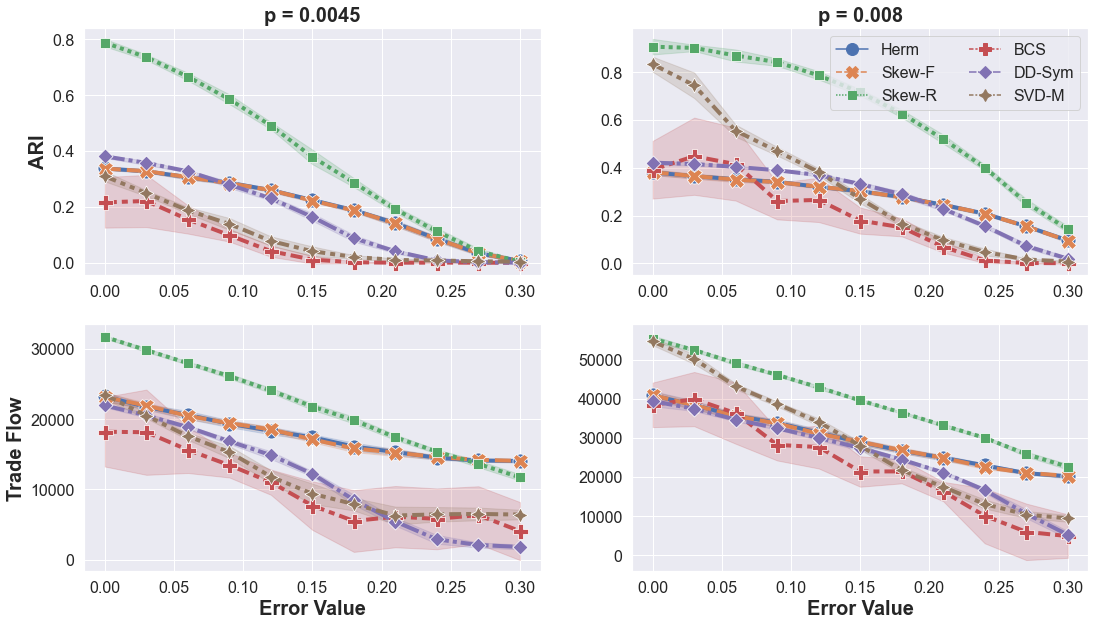}
\caption{ARI and TopTF results for the DAG DSBM experiments.}
  \label{fig:ari_uptri2}
\end{figure}

\begin{figure}
\centering
\includegraphics[width=1\textwidth]{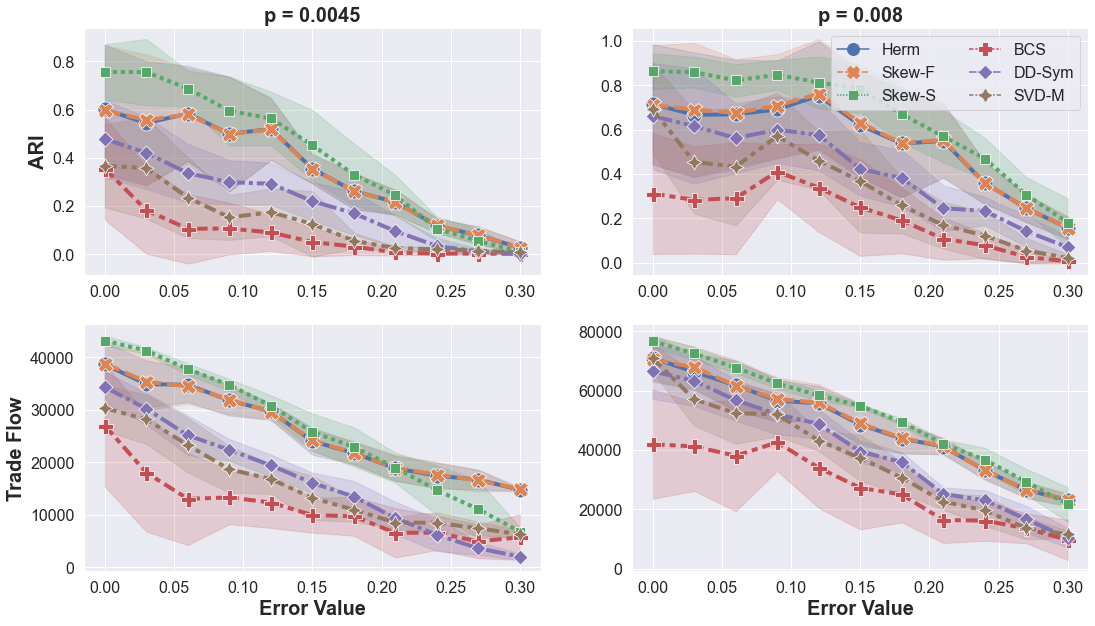}
  \caption{ARI results for the CMG DSBM experiments.}
  \label{fig:ari_cmdsbm}
\end{figure}

\begin{itemize}
\item {\it Circulant DSBM.}
\label{sec:circ_dsbm}
Here, the matrix $\M{F}$ is circulant.
This specific meta-graph model has received attention in recent work \cite{Cucuringu2019,Laenen_neruips,JCN_BCS} because it is a natural pattern of interest, and because it affords tools from the spectral theory of (block) circulant matrices, as well as Perron-Frobenius theory. 
Figure \ref{fig:ari_cyclic} present the results. The best performing algorithms are 
Hermitian Clustering and the two Skew-Symmetric Clustering algorithms.
We note for the Skew-R algorithm we set $l=1$, and for TopTF computation we set $c = k$, as this is the number of meaningful cuts expected to be found.
Lastly, we observe for extreme sparsity value, $p=0.0045$, using a single pair of singular vectors, the equivalent of using a single complex eigenvector, is better in terms of ARI and TopTF.

\item {\it Directed Acyclic DSBM.}
Here, the meta-graph resembles a Directed Acyclic Graph (DAG). This case is motivated by the fact that matrices $\M{F}$ constructed from DAGs have nonzero $\theta$-distinguishing images -- a requirement necessary for graphs generated by DSBM to statistically recoverable; see \cite{Cucuringu2019} for more. 
For our experiments we choose the DAG where the matrix $\M{F}_{uv} = \mu$ if $v=u+1$ or $v=u+2$, $\M{F}_{uv} = 1 - \mu$ if $v = u-1$ or $v = u-2$, and $\M{F}_{uv} = 1/2$ otherwise.
That is, the meta-graph is characterized by the first two lower and upper diagonals of $\M{F}$. 
Figure \ref{fig:ari_uptri2} present the results. 
When computing TopTF we set $c = 2(k-1)$, and again set $l=1$ for the Skew-R algorithm.
Results for this model are quite good for the SVD-R algorithm.
As was the case for Circulant DSBM, the choice of $l$ significantly impacts the results.
It is worth noting the meta-graph for the circulant case is strongly connected whereas in the DAG case it is only weakly connected.

\item {\it Complete Meta-Graph DSBM.}
In the CMG model \cite{Cucuringu2019}, $\M{F}$ is generated by randomly orienting the flows between clusters, and setting all entries of $\M{F}$, except for the diagonal, to either $\mu$ or $1-\mu$.
Figure \ref{fig:ari_cmdsbm} present the results. 
Skew-S significantly outperforms Herm and Skew-F in ARI, and slightly in terms of TopTF scores.
This demonstrates a static choice of the number of eigen or singular vectors, $l$, is not the most effective technique. 
Moreover, this suggests optimal choice of $l$ depends on the meta-graph pattern considered, rather than simply a function of the number of clusters. 
Lastly we note that the variance for these experiments is quite high. 
This is likely due to the fact that at each $\mu$ value ARI or TopTF scores from graphs with different $\M{F}$'s are being averaged.
As can be observed from the Circulant and DAG DSBM experiments, different patterns in $\M{F}$ exhibit different behaviors in ARI and TopTF scores.
\end{itemize}

We now consider timing results for the DSBM experiments. Summarized in Table \ref{tab:timings_table_p8}, timings are split into 3 parts:
\begin{enumerate}
    \item  Setup: the time spent forming the appropriate matrix representation, for example $\SM = \M{M} - \M{M}^\Tra$ or $\DDS = \M{M}^\Tra\M{M} + \M{M}\M{M}^\Tra$.
    \item Embedding (Emb.): the time spent selecting and computing the appropriate eigenvectors or singular vectors.
    \item $K$-means: time spent running the $k$-means algorithm.
\end{enumerate}

The last two columns present the speed up relative to Hermitian Clustering and the embedding dimension $k$-means is applied to, respectively. Run times for each $p$ value are averages of runs on 100 different graphs generated from the Circulant DSBM with each algorithm run on each graph 10 times.
We note all aglorithms were run on a computer with a 2.3 GHz Quad-Core Intel Core i7 processor and 32GB memory, with matrices stored using MATLAB's sparse matrix format, and utilizing MATLAB's kmeans(), eigs() and svds() functions. 
MATLAB was given access to all 4 CPUs during the experiments.

We observe the Hermitian Clustering algorithm runs much slower than the SVD based algorithms.
This primarly stems from running $k$-means on a dense $\numv \times \numv$ matrix, as was discussed in Section \ref{SS_discussion}.
For $p=0.008$ we observe speeds ups relative to Herm of about $ 13\times$ for both Skew-Symmetric Clustering variants, $5.3\times$ for BCS, $9\times$ for DD-Sym, and $16.5 \times$ for SVD-M.
Interestingly, we note the DD-Sym algorithm requires relatively more time in both Setup and Embedding as $p$ increases.
This is because the algorithm computes $\M{S}$ which becomes increasingly dense as $p$ increases due to the products $\M{M}\M{M}^\Tra$ and $\M{M}^\Tra\M{M}$.
This not only makes the computation of $\M{S}$ more expensive but also the computation of its eigenvectors.
Timing results for the other values of $p$ can be found in the Appendix, Section \ref{app_timing}.

\begin{table}
\centering
\begin{tabular}{l||c|c|c|c|c}%
\bfseries Alg. & \bfseries Setup & \bfseries 
Emb. & \bfseries Kmeans & \bfseries SpeedUp & \bfseries E. Dim.
\begin{filecontents*}{CSV_Data/p8_e.csv}
Alg,Setup,Embedding,Kmeans,SpeedUp, Dim
Herm,2.6,339.0,3510.5,1.0, $\numv \times \numv$
Skew-F,1.5,280.0,14.9,13.0, $\numv \times l$
Skew-R,1.8,286.3,12.1,12.8, $\numv \times 1$
BCS,0.8,704.1,16.7,5.3, $\numv \times 2\lfloor \frac{k}{2} \rfloor$
DD-Sym,63.7,342.7,17.9,9.1, $\numv \times k$
SVD-M,0.0,209.1,24.0,16.5, $\numv \times 2k$
\end{filecontents*}
\csvreader[head to column names]{CSV_Data/p8_e.csv}{}
{\\ \Alg & \Setup & \Embedding & \Kmeans & \SpeedUp & \Dim}
\end{tabular}
\caption{Timings, in milliseconds, for Cyclic DSBM experiments with $p=0.008$. Values are averages over 100 runs on 10 different graphs (10 runs per graph).}
\label{tab:timings_table_p8}
\end{table}

Summarizing the DSBM experiments, the Skew-Symmetric Clustering algorithms tend to perform the best, along with Hermitian Clustering, in terms of TopTF and ARI.
This is congruent with Proposition \ref{edm_prop} and results from \cite{Cucuringu2019}.
Additionally, the Skew-Symmetric Clustering algorithm is cheap in terms of storage and computational requirements while the Hermitian Clustering algorithm is not.
Skew-Symmetric Clustering's run time is among the lowest with the exception of the \ref{alg:svd_clustering} algorithm.
Lastly, we provided empirical evidence that the choice of $l$ often has a significant impact on the performance of the algorithm.
\subsection{Experiments on Real Data}
\label{sec:exp_real}
\begin{figure}
  \begin{minipage}[c]{0.6\textwidth}
    \includegraphics[width=\textwidth]{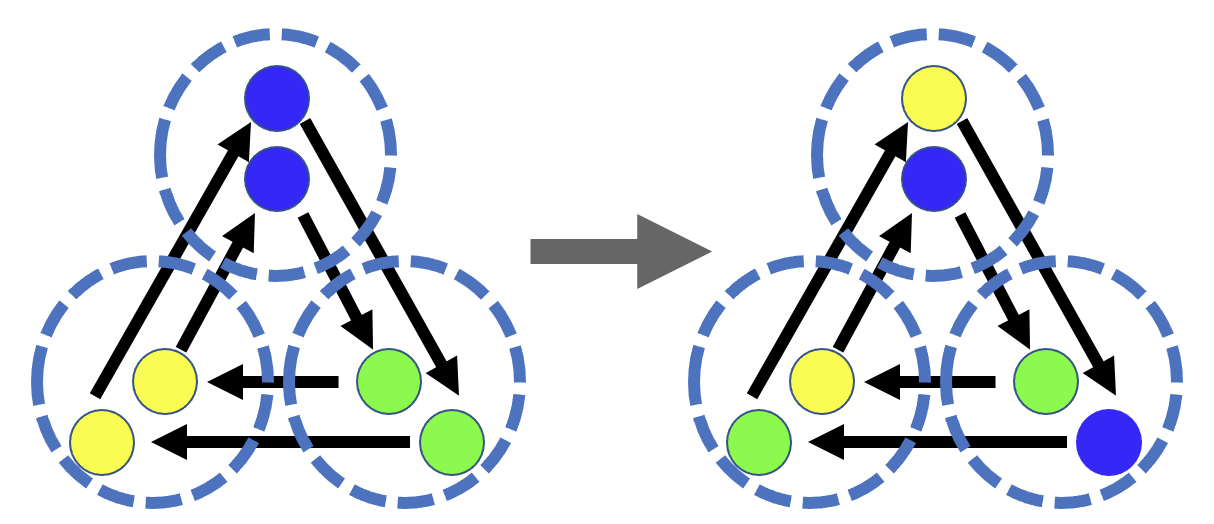}
  \end{minipage}\hfill
  \begin{minipage}[c]{0.39\textwidth}
    \caption{Two disjoint, oriented 3-cycles under two clusterings, indicated by the dotted lines.
    On the left, vertices of the same color are clustered, whereas on the right the outer cycle has been `rotated' clockwise by one position. 
    Both clusterings have equal TF values.
    } \label{fig:rot_fig}
  \end{minipage}
\end{figure}

Next, we test the performance of our algorithms on several food web datasets. Before proceeding, we discuss several practical considerations that are especially important when working with real data: connectedness and normalization. With regard to the former, similar to how undirected graph clustering algorithms typically require or enforce the input graph be connected, we ensure all digraphs we consider are weakly connected. This is for several reasons. First, disconnected components may cause issues with solution interpretation with flow-based clustering. For instance, Figure \ref{fig:rot_fig} provides an example illustrating how disconnectedness may yield multiple, trivially-valid clusterings. In general, however, we do not require strong connectivity. We note this distinction is important for certain algorithms. For instance the Block Cyclic Spectral Clustering algorithm relies on Perron-Frobenius theory to find flow-based clusterings, which requires the adjacency matrix $\M{M}$ be irreducible and hence the digraph strongly connected. Indeed, when the input is not strongly connected, the BCS algorithm regularizes $\M{M}$ to enforce strong connectivity; see \cite{JCN_BCS} for more. 

In the subsequent experiments, we utilize several normalizations, which can often improve results on real-world data. We note normalization is unnecessary for DSBM experiments, since vertices in the same clusters of the DSBM have the same expected value of in and out degree.
We consider two standard normalizations from spectral clustering that were utilized in \cite{Cucuringu2019}: defining $\M{D}_{uu} = \sum_{v}|\M{H}_{uv}|$, they are the symmetric normalization $\M{H}_{sym} = \M{D}^{-1/2}\M{H}\M{D}^{-1/2}$ and the random walk normalization $\M{H}_{rw} = \M{D}^{-1}\M{H}$.
One may analogously normalize $\M{K}$ as $\M{K}_{sym}$ and $\M{K}_{rw}$.
Based on prior results and recommendations \cite{Cucuringu2019, Luxburg2007}, we make use of the random walk normalization, $\M{H}_{rw}$ or $\M{K}_{rw}$, which we denoted by appending RW to the algorithm name, e.g. Herm-RW.
More discussion relating to normalization can be found in the Appendix, Section \ref{app:normalization}.


\paragraph{Florida Bay Food Web}
\begin{figure}[t]
\centering
     \includegraphics[width=0.95\linewidth]{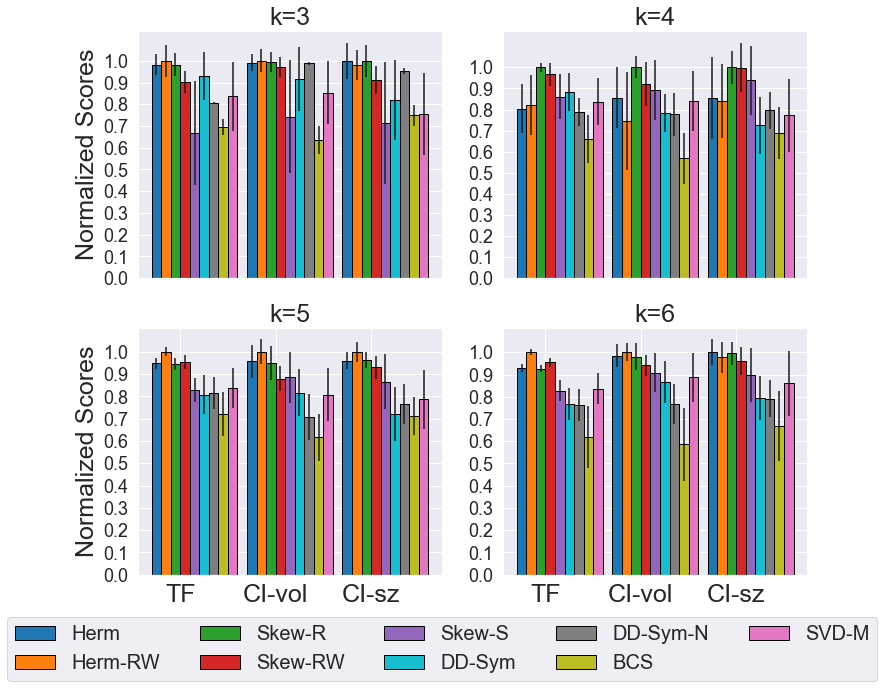}
  \caption{Cut scores on the FBFW for $k=\{ 3,4,5,6\}$. Bars corresponding to each metric are normalized by the highest achieved mean score so that all mean scores are between 0 and 1. Each bar's height is the mean over 100 runs and error bars give 1 standard deviation in each direction.}
  \label{fig:fb_barplts}
\end{figure}

\begin{figure}[t]
\centering
     \includegraphics[width=0.8\linewidth]{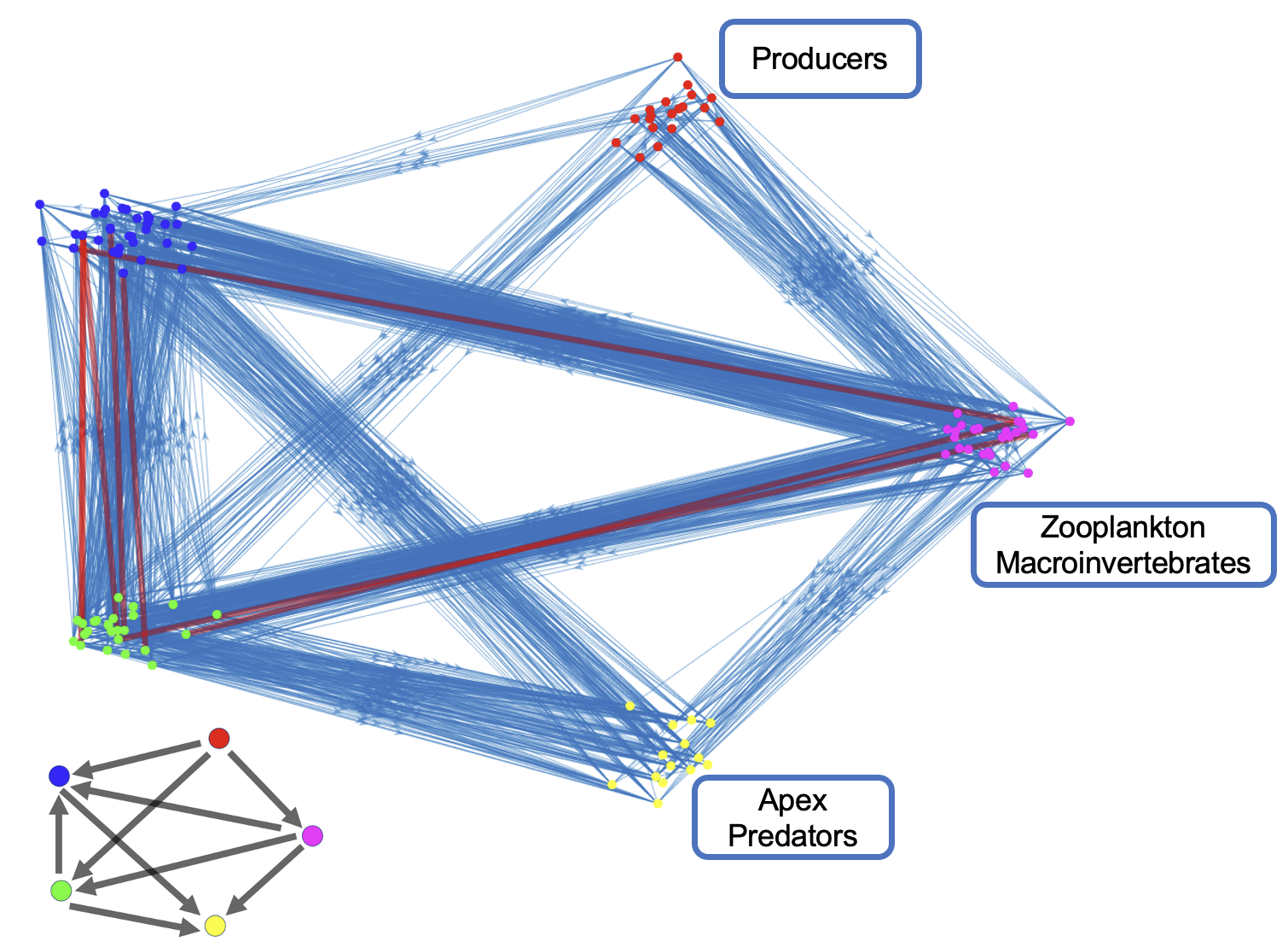}
  \caption{Plot of vertices and edges for the Skew-RW clustering with $k=5$ clusters on the 118 vertex subgraph from Li and Milekovic \cite{pan_li_neurips}. Edges oriented against the cluster hierarchy are in red and all other edges are in blue. The cluster hierarchy is red $\rightarrow$ magenta $\rightarrow$ green $\rightarrow$ blue $\rightarrow$ yellow. The matching bottom left graph, gives the cluster level orientations of edges. We add labels to 3 clusters which exhibit consistent within cluster species labels. Some parameters of the run include $l=1$ and the shown cluster is the best TopTF scorer over 100 runs. Table \ref{tab:FBFW_table} gives a full list of species and their clusters as show in the figure.}
  \label{fig:fb_k5_orientation_graph}
\end{figure}

The Florida Bay Food Web (FBFW)\footnote{http://vlado.fmf.uni-lj.si/pub/networks/data/bio/foodweb/foodweb.htm} is a data set containing information about carbon exchange between species in the South Florida Ecosystems.
In this digraph an edge $u \rightarrow v$ might mean species $v$ eats species $u$.
The graph contains 128 vertices and 2106 edges.
We treat the graph as unweighted.

This data set has been analyzed quite extensively \cite{DBLP:journals/corr/BensonGL16a,pan_li_neurips,JCN_BCS}, and is worth reviewing prior analyses to place ours in context.
Benson et al. \cite{DBLP:journals/corr/BensonGL16a} consider this dataset to demonstrate the effectiveness of their spectral motif clustering algorithm.
Their clustering results separate out a number of interesting within-cluster dynamics.
Li and Milenkovic \cite{pan_li_neurips} also used a spectral motif clustering approach in the context of their proposed inhomogenous hypergraph clustering problem.
Their choice of motif resulted in finding 5 clusters where most of the edges are oriented between clusters, thus revealing the hierarchical structure present in the graph.
The clusters in this hierarchical structure are roughly interpretable as the trophic levels, or cluster level predator-prey relationships.
Due to the nature of motif clustering Li and Milenkovic pruned the network of 10 vertices corresponding to `singleton' clusters (manatee, kingfisher, hawksbill turtle, etc.) and detritus species.
The resulting reduced network consists of 118 vertices and 1714 edges.
Impressively, in their clustering only 5 edges are oriented from higher to lower clusters in the found cluster hierarchy.
We directly compare against Algorithm \ref{alg:bcs_clustering} \cite{JCN_BCS} and so do not discuss their results in text.

Our method is able to uncover a similar hierarchical structure.
Skew-Symmetric Clustering with random walk normalization, Skew-RW, and $l=1$ gives the best results out of all methods run on this graph in terms of TopTF.
All of the methods previously compared against via the DSBM are able to operate on the full 128 vertices of the original graph.
However, for direct comparison we generate Li and Milenkovic's subgraph and compare their reported clustering versus that returned by Skew-RW.
Li and Milenkovic's clustering yields a TopTF score of 1536 which means that $\approx 89.6$\% of the 1714 edges are oriented between clusters according the to cluster hierarchy.
Skew-RW yields a maximum TopTF score of 1587, thus orienting about $\approx 92.6$\% of edges between clusters according to the hierarchy.
When computing TopTF we take all cluster-cluster relations into account (setting $c$ in Eqn. \ref{eq:toptf}).

The main difference between the Skew-RW clustering and that of Li and Milekovic is that our clustering has fewer within cluster edges, which of course do not contribute to the TopTF score.
The most prominent example of this is that our clustering places the algae and seagrass species in the cluster lowest in the hierarchy while Li and Milekovic places them in the second lowest.
In some sense, our clustering may be more intuitive.
For example the species Drift Algae and Epiphytes, also an algae, have no incoming edges in the reduced digraph.
While placing these species in the second lowest cluster does not introduce any edges oriented against the hierarchy it does result in more within cluster edges, thus lowering the overall TopTF score.
The clusters at the top of both hierarchies are identical.
This top cluster contains species such as sharks and dolphins.
We also note that our clustering orients 8 edges against the hierarchy while Li and Milekovic's clustering has only 5 such edges.
Our clustering is visualized in Figure \ref{fig:fb_k5_orientation_graph} and Table \ref{tab:FBFW_table} list all species, their names, and cluster assignment.

Figure \ref{fig:fb_barplts} presents cut score results for $k=\{3,4,5,6\}$.
The methods which utilize the matrices $\M{H}$ and $\M{K}$ are generally most successful across all 3 computed cut metrics.
Observe that Skew-R ($l=1$) and Skew-RW ($l=1$) are consistently two of the best performing algorithms in terms of cut scores.
Running these algorithms with $k>6$ did not increase, and often decreased, TopTF scores. DD-Sym-N refers to a normalized version of DD-Sym, see Satuluri and Parthasarathy \cite{DD_sym_citation} for details.
Since when $k=2$ the TF maximization problem is solveable we can compare the scores of the algorithms to the true max of Eqn. \ref{FI_maximization}.
The Skew-RW algorithm achieves the highest TF score of $1105$ which is about 95\% of the true max TF of $1163$.

\begin{figure}[t!]
\centering
\begin{subfigure}[t]{0.48\linewidth}
\centering
     \includegraphics[width=1\linewidth]{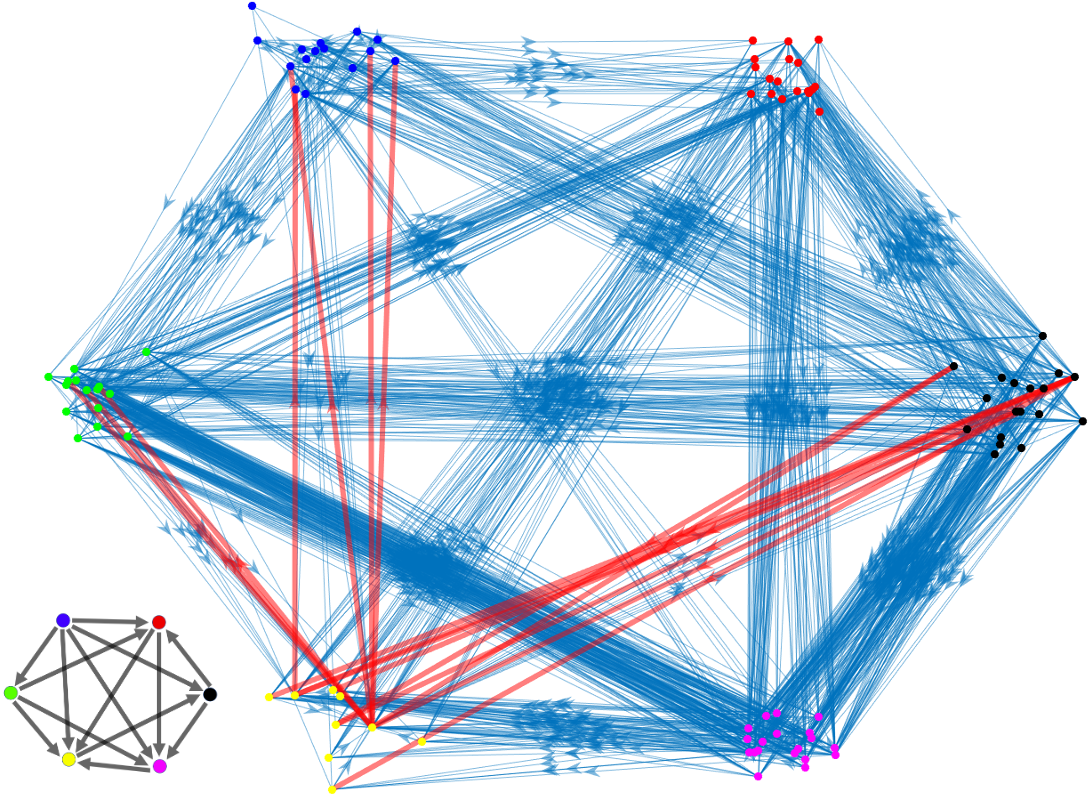}
  \caption{Visualization of the clustering output by Hermitian Clustering with random walk normalization on the 
  Mangrove (wet season) data set. There are 16 edges, indicated in bold and red, which actively detract from the TF score of 1104.}
  \label{fig:mang_plt1}
\end{subfigure}
\hspace{0.0\linewidth}
\begin{subfigure}[t]{0.48\linewidth}
\centering
     \includegraphics[width=1\linewidth]{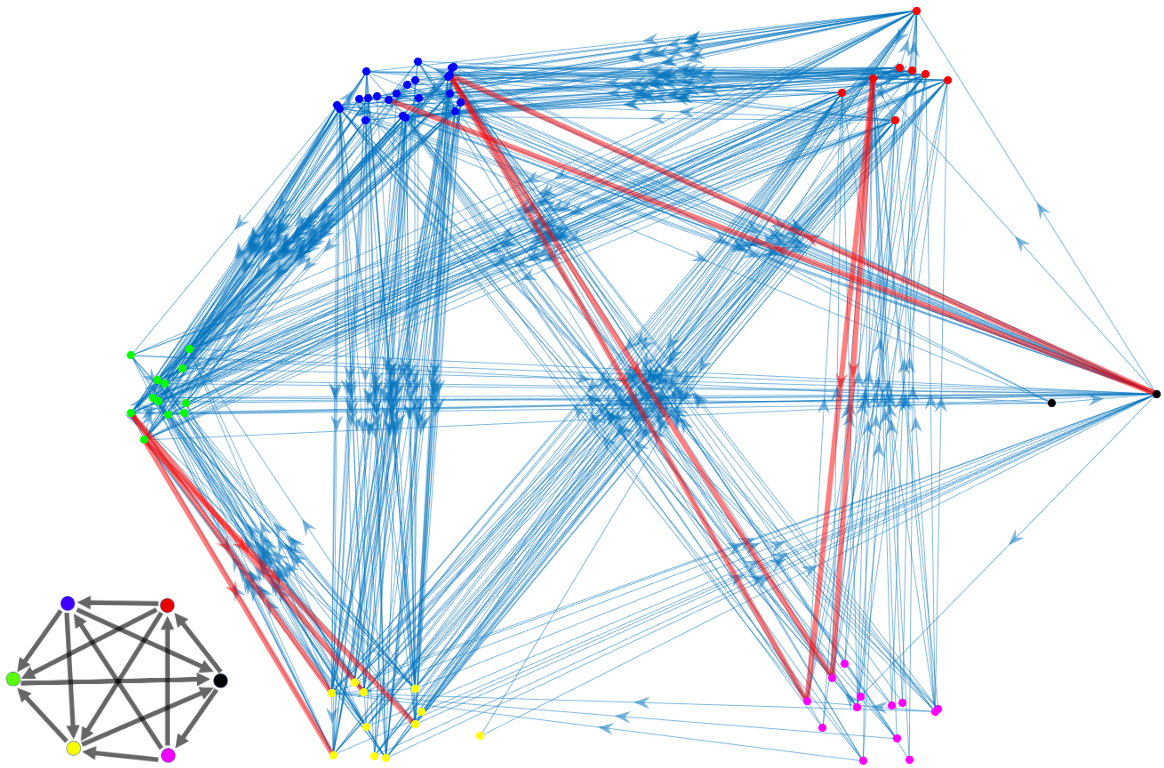}
  \caption{Visualization of the clustering output by Hermitan Clustering with random walk normalization on the Cypress Dry Season data set. There are 10 edges oriented against the majority flows. Approximately 89\% of the edges contribute to the TF score of 472.}
  \label{fig:cy_plt1}
\end{subfigure}
\end{figure}

\paragraph{Other Food Webs}
We briefly present results for two other food web data sets: Mangrove Wet Season and Cypress Dry Season, which are originally Pajek datasets\footnote{http://vlado.fmf.uni-lj.si/pub/networks/data/bio/foodweb/foodweb.htm}.
Visualizations of output clusterings that achieve the highest TopTF score for the Hermitian Clustering algorithm with random walk normalization (Herm-RW) can be seen in Figure \ref{fig:mang_plt1} and Figure \ref{fig:cy_plt1}.
The Herm-RW algorithm gives on average the highest TopTF scores for both of these graphs but the Skew algorithm is also able to recover the clustering which yields the highest found TopTF score.
Again we observe that the method is able to uncover a clustering structure, with $k=6$, that yields a high TopTF score and appears to reveal a cluster level hierarchical structure. 

\section{Conclusion}
We've explored the role of complex-valued adjacency matrices for finding imbalanced cuts in directed graphs. 
Through a careful analysis of algebraic relationships we show that real valued representation and algorithms which use real arithmetic are not only possible but advantageous.
Our algorithm, Skew-Symmetric Clustering, is asymptotically faster and requires less memory than the existing state of the art method.
It also has a natural connection to a simple metric which captures the spirit of imbalanced cuts.
We demonstrate the algorithms ability to find meaningful patterns in real world data and outperform related methods on graphs generated form the Directed Stochastic Block Model.

In a broader sense we hope that this work will encourage careful consideration of the role of complex-valued representations for graphs.
While our work primarily focuses on algorithmic drawbacks of using complex-valued representations, there are advantages for considering such matrices. 
For example Cucuringu et al. \cite{Cucuringu2019} use the Davis Kahan Theorem \cite{Davis_kahan} for Hermitian matrices in their analysis and Laenen and Sun \cite{Laenen_neruips} use the fact that Hermitian matrices are subject to the min-max theorem.
Some limitations of this work include focusing on a single, specific complex-valued digraph matrix limited to oriented graphs, the non-generality of the relaxation argument with respect to $k$, and a further results on larger, real-world graphs.


\section*{Acknowledgments}
Information Release PNNL-SA-162707.

\appendix
\section{Normalization}
\label{app:normalization}
In Section \ref{sec:exp_real} various normalizations of the algorithms based on $\M{H}$ and $\M{S}$ are considered.
The experiments show that normalization is clearly advantageous when dealing with real world data.
The normalizations considered are $\M{H}_{rw} = \M{D}^{-1}\M{H}$ and $\M{K}_{rw} = \M{D}^{-1}\M{K}$.
Because $\M{H}_{rw}$ and $\M{K}_{rw}$ are not normal, though they are still diagonalizeable because they are similar to diagonalizeable matrices, we cannot relate the SVD of $\M{K}_{rw}$ to its EVD.
Therefore the analysis presented in Section \ref{sec:CAM_sec} is not applicable or easy generalized.
However, the arguments in Section \ref{sec:CAM_sec} hold for $\M{H}_{sym}$ and $\M{K}_{sym}$ as both matrices are normal.

\section{Timing Data}
Tables showing timings for Circulant DSBM experiments with $p = [0.0045,0.005,0.006]$.
\label{app_timing}
\begin{table}[h!]
\centering
\begin{subtable}[h]{\textwidth}\centering
\captionsetup{justification=centering}
\begin{tabular}{l||c|c|c|c}%
\bfseries Alg. & \bfseries Setup & \bfseries 
Emb. & \bfseries Kmeans & \bfseries SpeedUp
\begin{filecontents*}{CSV_Data/p45.csv}
Alg,Setup,Embedding,Kmeans,SpeedUp
Herm,1.3,357.5,5406.6,1.0
Skew-F,1.0,248.2,26.5,20.9
Skew-R,1.1,253.8,17.8,21.1
BCS,0.9,867.1,18.4,6.5
DD-Sym,33.1,160.6,25.2,26.3
SVD-M,0.0,203.1,29.0,24.8
\end{filecontents*}
\csvreader[head to column names]{CSV_Data/p45.csv}{}
{\\ \Alg & \Setup & \Embedding & \Kmeans & \SpeedUp}
\end{tabular}
\caption{Timings for $p=0.0045$}
\label{tab:timing_p45}
\end{subtable}
\\
\begin{subtable}[h]{\textwidth}\centering
\begin{tabular}{l||c|c|c|c}%
\bfseries Alg. & \bfseries Setup & \bfseries 
Emb. & \bfseries Kmeans & \bfseries SpeedUp
\begin{filecontents*}{CSV_Data/p5.csv}
Alg,Setup,Embedding,Kmeans,SpeedUp
Herm,1.2,322.0,4934.0,1.0
Skew-F,1.0,217.0,22.0,21.9
Skew-R,1.1,221.7,15.5,22.1
BCS,0.7,713.5,16.1,7.2
DD-Sym,33.3,173.9,22.0,22.9
SVD-M,0.0,184.8,26.4,24.9
\end{filecontents*}
\csvreader[head to column names]{CSV_Data/p5.csv}{}
{\\ \Alg & \Setup & \Embedding & \Kmeans & \SpeedUp}
\end{tabular}
\caption{Timings for $p=0.005$}
\label{tab:timing_p5}
\end{subtable}
\\
\begin{subtable}[h]{\textwidth}\centering
\begin{tabular}{l||c|c|c|c}%
\bfseries Alg. & \bfseries Setup & \bfseries 
Emb. & \bfseries Kmeans & \bfseries SpeedUp
\begin{filecontents*}{CSV_Data/p6.csv}
Alg,Setup,Embedding,Kmeans,SpeedUp
Herm,1.5,330.2,4433.3,1.0
Skew-F,1.1,230.8,19.1,19.0
Skew-R,1.3,237.2,14.3,18.8
BCS,0.7,717.2,15.7,6.5
DD-Sym,41.7,240.2,21.0,15.7
SVD-M,0.0,199.9,26.5,21.0
\end{filecontents*}
\csvreader[head to column names]{CSV_Data/p6.csv}{}
{\\ \Alg & \Setup & \Embedding & \Kmeans & \SpeedUp}
\end{tabular}
\caption{Timings for $p=0.006$}
\label{tab:timing_p6}
\end{subtable}
\end{table}

\clearpage
\section{Florida Bay Food Web Clustering}
\label{app_FBFW}

\begin{table}[b!]
\centering
\scalebox{0.57}{
\centering
\begin{tabular}[t]{l|l|l|}%
\bfseries Name & \bfseries 
Group & \bfseries Cluster
\begin{filecontents*}{CSV_Data/SkewRW_PanLi_graph1.csv}
nodeid,name,group,colorlab
0,2um Spherical Phytoplankt,x,red
1,Synedococcus,x,red
2,Oscillatoria,x,red
3,Small Diatoms (<20um),x,red
4,Big Diatoms (>20um),x,red
5,Dinoflagellates,x,red
6,Other Phytoplankton,x,red
7,Benthic Phytoplankton,Demersal Producer,red
8,Thalassia,Seagrass Producer,red
9,Halodule,Seagrass Producer,red
10,Syringodium,Seagrass Producer,red
12,Drift Algae,Algae Producer,red
13,Epiphytes,Algae Producer,red
14,Free Bacteria,Microbial Microfauna,red
15,Water Flagellates,Microbial Microfauna,red
16,Water Cilitaes,Microbial Microfauna,red
23,Benthic Flagellates,Sediment Organism Microfauna,red
24,Benthic Ciliates,Sediment Organism Microfauna,red
25,Meiofauna,Sediment Organism Microfauna,red
28,Other Cnidaridae,Macroinvertebrates,blue
51,Stone Crab,x,blue
53,Rays,x,blue
55,Bonefish,x,blue
59,Lizardfish,Benthic Fishes,blue
60,Catfish,Benthic Fishes,blue
61,Eels,Demersal Fishes,blue
63,Brotalus,Demersal Fishes,blue
65,Needlefish,Pelagic Fishes,blue
69,Snook,x,blue
76,Jacks,x,blue
77,Pompano,x,blue
78,Other Snapper,x,blue
79,Gray Snapper,x,blue
81,Grunt,x,blue
82,Porgy,x,blue
84,Scianids,x,blue
85,Spotted Seatrout,x,blue
86,Red Drum,x,blue
87,Spadefish,x,blue
95,Flatfish,Benthic Fishes,blue
96,Filefishes,x,blue
97,Puffer,x,blue
98,Other Pelagic Fishes,Pelagic Fishes,blue
105,Small Herons + Egrets,x,blue
106,Ibis,x,blue
107,Roseate Spoonbill,x,blue
108,Herbivorous Ducks,x,blue
109,Omnivorous Ducks,x,blue
112,Gruiformes,x,blue
113,Small Shorebirds,x,blue
114,Gulls + Terns,x,blue
117,Loggerhead Turtle,x,blue
118,Green Turtle,x,blue
\end{filecontents*}
\csvreader[head to column names]{CSV_Data/SkewRW_PanLi_graph1.csv}{}
{\\ \name & \group & \colorlab}
\end{tabular}
\begin{tabular}[t]{l|l|l}%
\bfseries Name & \bfseries 
Group & \bfseries Cluster
\begin{filecontents*}{CSV_Data/SkewRW_PanLi_graph2.csv}
nodeid,name,group,colorlab
27,Coral,x,green
29,Echinoderma,Macroinvertebrates,green
46,Lobster,x,green
49,Predatory Crabs,Macroinvertebrates,green
50,Callinectus sapidus,Macroinvertebrates,green
56,Sardines,Pelagic Fishes,green
57,Anchovy,Pelagic Fishes,green
58,Bay Anchovy,Pelagic Fishes,green
62,Toadfish,Benthic Fishes,green
64,Halfbeaks,Pelagic Fishes,green
66,Other Killifish,x,green
67,Goldspotted killifish,Demersal Fishes,green
68,Rainwater killifish,Demersal Fishes,green
71,Silverside,Pelagic Fishes,green
72,Other Horsefish,x,green
73,Gulf Pipefish,x,green
74,Dwarf Seahorse,x,green
80,Mojarra,x,green
83,Pinfish,x,green
88,Parrotfish,x,green
90,Mullet,Pelagic Fishes,green
92,Blennies,Benthic Fishes,green
93,Code Goby,Benthic Fishes,green
94,Clown Goby,Benthic Fishes,green
99,Other Demersal Fishes,Demersal Fishes,green
52,Sharks,x,yellow
54,Tarpon,x,yellow
75,Grouper,x,yellow
89,Mackerel,x,yellow
91,Barracuda,x,yellow
100,Loon,x,yellow
101,Greeb,x,yellow
102,Pelican,x,yellow
103,Comorant,x,yellow
104,Big Herons + Egrets,x,yellow
110,Predatory Ducks,x,yellow
111,Raptors,x,yellow
116,Crocodiles,x,yellow
120,Dolphin,x,yellow
17,Acartia Tonsa,Zooplankton Microfauna,purple
18,Oithona nana,Zooplankton Microfauna,purple
19,Paracalanus,Zooplankton Microfauna,purple
20,Other Copepoda,Zooplankton Microfauna,purple
21,Meroplankton,Zooplankton Microfauna,purple
22,Other Zooplankton,Zooplankton Microfauna,purple
26,Sponges,Macroinvertebrates,purple
30,Bivalves,Macroinvertebrates,purple
31,Detritivorous Gastropods,Macroinvertebrates,purple
32,Epiphytic Gastropods,x,purple
33,Predatory Gastropods,Macroinvertebrates,purple
34,Detritivorous Polychaetes,Macroinvertebrates,purple
35,Predatory Polychaetes,Macroinvertebrates,purple
36,Suspension Feeding Polych,Macroinvertebrates,purple
37,Macrobenthos,Macroinvertebrates,purple
38,Benthic Crustaceans,Macroinvertebrates,purple
39,Detritivorous Amphipods,Macroinvertebrates,purple
40,Herbivorous Amphipods,Macroinvertebrates,purple
41,Isopods,Macroinvertebrates,purple
42,Herbivorous Shrimp,Macroinvertebrates,purple
43,Predatory Shrimp,Macroinvertebrates,purple
44,Pink Shrimp,Macroinvertebrates,purple
45,Thor Floridanus,x,purple
47,Detritivorous Crabs,Macroinvertebrates,purple
48,Omnivorous Crabs,Macroinvertebrates,purple
70,Sailfin Molly,x,purple
\end{filecontents*}
\csvreader[head to column names]{CSV_Data/SkewRW_PanLi_graph2.csv}{}
{\\ \name & \group & \colorlab}
\end{tabular}
}
\caption{Our results on the FBFW. Columns are names, higher level species classification if available, and clustering according to Figure \ref{fig:fb_k5_orientation_graph}. A value of 'x' means there is no given label or name.}
\label{tab:FBFW_table}
\end{table}
\clearpage


\bibliographystyle{siamplain}
\bibliography{references}

\begin{thebibliography}{10}

\bibitem{DBLP:journals/corr/BensonGL16a}
{\sc A.~R. Benson, D.~F. Gleich, and J.~Leskovec}, {\em Higher-order
  organization of complex networks}, CoRR, abs/1612.08447 (2016),
  \url{http://arxiv.org/abs/1612.08447},
  \url{https://arxiv.org/abs/1612.08447}.

\bibitem{Cucuringu2019}
{\sc M.~Cucuringu, H.~Li, H.~Sun, and L.~Zanetti}, {\em Hermitian matrices for
  clustering directed graphs: insights and applications}, arXiv:1908.02096,
  (2019), \url{https://arxiv.org/abs/1908.02096v1}.

\bibitem{Davis_kahan}
{\sc C.~Davis and W.~M. Kahan}, {\em The rotation of eigenvectors by a
  perturbation. iii}, SIAM Journal on Numerical Analysis, 7 (1970), pp.~1--46,
  \url{https://doi.org/10.1137/0707001}, \url{https://doi.org/10.1137/0707001},
  \url{https://arxiv.org/abs/https://doi.org/10.1137/0707001}.

\bibitem{EDM_Theory}
{\sc I.~Dokmanic, R.~Parhizkar, J.~Ranieri, and M.~Vetterli}, {\em Euclidean
  distance matrices: Essential theory, algorithms and applications}, Signal
  Processing Magazine, IEEE, 32 (2015), pp.~12--30,
  \url{https://doi.org/10.1109/MSP.2015.2398954}.

\bibitem{furutani2019graph}
{\sc S.~Furutani, T.~Shibahara, M.~Akiyama, K.~Hato, and M.~Aida}, {\em Graph
  signal processing for directed graphs based on the hermitian laplacian.}, in
  ECML/PKDD (1), 2019, pp.~447--463.

\bibitem{ARI_cit}
{\sc A.~Gates and Y.-Y. Ahn}, {\em The impact of random models on clustering
  similarity}, Journal of Machine Learning Research, 18 (2017).

\bibitem{gleich2006hierarchical}
{\sc D.~Gleich}, {\em Hierarchical directed spectral graph partitioning},
  Information Networks,  (2006).

\bibitem{GandVL}
{\sc G.~H. Golub and C.~F. van Loan}, {\em Matrix Computations}, JHU Press,
  fourth~ed., 2013,
  \url{http://www.cs.cornell.edu/cv/GVL4/golubandvanloan.htm}.

\bibitem{greub1967linear}
{\sc W.~Greub}, {\em Linear algebra}, Springer, Berlin,New York, third~ed.,
  1967.

\bibitem{Guo_mixed_rep}
{\sc K.~Guo and B.~Mohar}, {\em Hermitian adjacency matrix of digraphs and
  mixed graphs}, Journal of Graph Theory, 85 (2017), pp.~217--248,
  \url{https://doi.org/https://doi.org/10.1002/jgt.22057},
  \url{https://onlinelibrary.wiley.com/doi/abs/10.1002/jgt.22057},
  \url{https://arxiv.org/abs/https://onlinelibrary.wiley.com/doi/pdf/10.1002/jgt.22057}.

\bibitem{Vemplala_spec_clust}
{\sc R.~Kannan, S.~Vempala, and A.~Vetta}, {\em On clusterings: Good, bad and
  spectral}, J. ACM, 51 (2004), p.~497–515,
  \url{https://doi.org/10.1145/990308.990313},
  \url{https://doi.org/10.1145/990308.990313}.

\bibitem{laenen_MS_2019}
{\sc S.~Laenen}, {\em Directed graph clustering using hermitian laplacians},
  master's thesis, 2019.

\bibitem{Laenen_neruips}
{\sc S.~{Laenen} and H.~{Sun}}, {\em {Higher-Order Spectral Clustering of
  Directed Graphs}}, arXiv e-prints,  (2020), arXiv:2011.05080,
  p.~arXiv:2011.05080, \url{https://arxiv.org/abs/2011.05080}.

\bibitem{pan_li_neurips}
{\sc P.~Li and O.~Milenkovic}, {\em Inhomogeneous hypergraph clustering with
  applications}, in Advances in Neural Information Processing Systems,
  I.~Guyon, U.~V. Luxburg, S.~Bengio, H.~Wallach, R.~Fergus, S.~Vishwanathan,
  and R.~Garnett, eds., vol.~30, Curran Associates, Inc., 2017,
  \url{https://proceedings.neurips.cc/paper/2017/file/a50abba8132a77191791390c3eb19fe7-Paper.pdf}.

\bibitem{liu2015hermitian}
{\sc J.~Liu and X.~Li}, {\em Hermitian-adjacency matrices and hermitian
  energies of mixed graphs}, Linear Algebra and its Applications, 466 (2015),
  pp.~182--207.

\bibitem{digraph_clust_survey}
{\sc F.~D. Malliaros and M.~Vazirgiannis}, {\em Clustering and community
  detection in directed networks: A survey}, Physics Reports, 533 (2013),
  pp.~95--142,
  \url{https://doi.org/https://doi.org/10.1016/j.physrep.2013.08.002},
  \url{https://www.sciencedirect.com/science/article/pii/S0370157313002822}.
\newblock Clustering and Community Detection in Directed Networks: A Survey.

\bibitem{MOHAR_w6}
{\sc B.~Mohar}, {\em A new kind of hermitian matrices for digraphs}, Linear
  Algebra and its Applications, 584 (2020), pp.~343--352,
  \url{https://doi.org/https://doi.org/10.1016/j.laa.2019.09.024},
  \url{https://www.sciencedirect.com/science/article/pii/S0024379519304136}.

\bibitem{andrew_Ng_spec_clust}
{\sc A.~Y. Ng, M.~I. Jordan, and Y.~Weiss}, {\em On spectral clustering:
  Analysis and an algorithm}, in Proceedings of the 14th International
  Conference on Neural Information Processing Systems: Natural and Synthetic,
  NIPS'01, Cambridge, MA, USA, 2001, MIT Press, pp.~849--856.

\bibitem{Peng_spec_clust}
{\sc R.~Peng, H.~Sun, and L.~Zanetti}, {\em Partitioning well-clustered graphs:
  Spectral clustering works!}, in Proceedings of The 28th Conference on
  Learning Theory, P.~Grünwald, E.~Hazan, and S.~Kale, eds., vol.~40 of
  Proceedings of Machine Learning Research, Paris, France, 03--06 Jul 2015,
  PMLR, pp.~1423--1455, \url{https://proceedings.mlr.press/v40/Peng15.html}.

\bibitem{DD_sym_citation}
{\sc V.~Satuluri and S.~Parthasarathy}, {\em Symmetrizations for clustering
  directed graphs}, in Proceedings of the 14th International Conference on
  Extending Database Technology, EDBT/ICDT '11, New York, NY, USA, 2011,
  Association for Computing Machinery, pp.~343--354,
  \url{https://doi.org/10.1145/1951365.1951407},
  \url{https://doi.org/10.1145/1951365.1951407}.

\bibitem{normalized_cuts_image_seg}
{\sc J.~Shi and J.~Malik}, {\em Normalized cuts and image segmentation}, IEEE
  Transactions on Pattern Analysis and Machine Intelligence, 22 (2000),
  pp.~888--905, \url{https://doi.org/10.1109/34.868688}.

\bibitem{svdm_paper}
{\sc D.~Sussman, M.~Tang, D.~Fishkind, and C.~Priebe}, {\em A consistent
  adjacency spectral embedding for stochastic blockmodel graphs}, Journal of
  the American Statistical Association, 107 (2011),
  \url{https://doi.org/10.1080/01621459.2012.699795}.

\bibitem{JCN_BCS}
{\sc H.~Van~Lierde, T.~W.~S. Chow, and J.-C. Delvenne}, {\em {Spectral
  clustering algorithms for the detection of clusters in block-cyclic and
  block-acyclic graphs}}, Journal of Complex Networks, 7 (2018), pp.~1--53,
  \url{https://doi.org/10.1093/comnet/cny011},
  \url{https://doi.org/10.1093/comnet/cny011},
  \url{https://arxiv.org/abs/https://academic.oup.com/comnet/article-pdf/7/1/1/27737475/cny011.pdf}.

\bibitem{Luxburg2007}
{\sc U.~von Luxburg}, {\em A tutorial on spectral clustering}, Statistics and
  Computing, 17 (2007), pp.~395--416,
  \url{https://doi.org/10.1007/s11222-007-9033-z},
  \url{http://dx.doi.org/10.1007/s11222-007-9033-z}.

\bibitem{zhang2021magnet}
{\sc X.~Zhang, Y.~He, N.~Brugnone, M.~Perlmutter, and M.~Hirn}, {\em Magnet: A
  neural network for directed graphs}, arXiv preprint arXiv:2102.11391,
  (2021).

\end{thebibliography}
\end{document}